\pdfoutput=1

\documentclass[journal]{IEEEtran}
\ifCLASSINFOpdf
  \usepackage[pdftex]{graphicx}
  \graphicspath{{./figs/}}
\else
\fi
\usepackage{amsmath}
\usepackage{amssymb} 

\usepackage{algpseudocode}

\newcommand{\eg}{{e.g. }}
\newcommand{\etal}{\textit{{et al.} }}

\usepackage{color,soul}

\begin{document}
%
\title{On-device Scalable Image-based Localization via Prioritized Cascade Search and Fast
One-Many RANSAC}
%
%
%

\author{Ngoc-Trung Tran,
        Dang-Khoa Le Tan,
        Anh-Dzung Doan,
        Thanh-Toan Do,
        Tuan-Anh Bui,
        Mengxuan Tan,
        Ngai-Man Cheung
\noindent \thanks{The authors are with the Singapore University of Technology and Design (SUTD), Singapore. Corresponding authors: ngoctrung\_tran@sutd.edu.sg, ngaiman\_cheung@sutd.edu.sg}

}

\maketitle

\begin{abstract}

We present the design of an entire on-device system for large-scale urban localization using images. The proposed design integrates compact image retrieval and 2D-3D correspondence search to estimate the location in extensive city regions. Our design is GPS agnostic and does not require network connection.  
In order to overcome the resource constraints of mobile devices, we 
propose a system design that leverages the scalability advantage of image retrieval and accuracy of 3D model-based localization. Furthermore, we propose a new hashing-based cascade search for fast computation of 2D-3D correspondences. In addition, we propose a new one-many RANSAC for accurate pose estimation. The new one-many RANSAC addresses the challenge of repetitive building structures (e.g. windows, balconies) in urban localization. Extensive experiments demonstrate that our 2D-3D correspondence search achieves state-of-the-art localization accuracy on multiple benchmark datasets. Furthermore, our experiments on a large Google Street View (GSV) image dataset show the potential of large-scale localization entirely on a typical mobile device.

\end{abstract}


%
\IEEEpeerreviewmaketitle

\section{Introduction}

\IEEEPARstart{E}{stimating} accurately the camera pose is a fundamental requirement of many applications, including robotics, augmented reality, autonomous vehicle navigation and location recognition. Usage of visual/image sensors (e.g., camera) is advantageous when developing such localization system because they provide rich information about the scene. While sensor data obtained from GPS (Global Positioning System), WiFi and Bluetooth can also be used, they have their limitations. 
The accuracy of GPS sensors is highly dependent on the surrounding environments. GPS-based localization would perform poorly in downtown areas and urban canyons, e.g., the localization error can be up to 30m or more \cite{chen-cvpr-2011}. Moreover, GPS information is often unavailable in indoor locations. Due to its sensitivity to magnetic disturbances,  GPS can also be denied/lost or easily hacked, and thus is not suitable for secure applications. While  localization systems using WiFi and Bluetooth can be considered, they are not always available in outdoor environments. Therefore, it is important to investigate image-based localization systems that  do not require GPS/Bluetooth/WiFi support.

State-of-the-art methods for image-based localization \cite{li-eccv-2012,svarm-cvpr-2014,sattler-pami-2016} leverage the  3D models of the  scene. These 3D  models are often pre-built from image datasets by using advanced Structure-from-Motion (SfM)  \cite{snavely-siggraph-2006}. These 3D model based localization methods are memory and computational intensive.  It is challenging to employ them on resource-constrained mobile devices \cite{zhou:2018}. 


The main goal of our work is to research a large-scale localization system that  runs entirely on a mobile device. We address following main challenges: constrained memory and  computational resources of a mobile device, requirements of high localization accuracy and extensive localization coverage.
Previous work has not addressed all these challenges in a single solution. In particular,  previous work has focused on improving accuracy \cite{li-eccv-2010,li-eccv-2012,svarm-cvpr-2014,sattler-pami-2016}. Other work has proposed systems  on mobile devices but they require client-server communication  due to high computational requirements \cite{arth-ismar-2011,ventura-tvcg-2014,middelberg-eccv-2014}. Some work has researched on-device systems but they cover only small areas due to memory usages \cite{lim-cvpr-2012}. 
To address all the challenges, 
our paper makes novel contributions in both system design and component image processing algorithms.


{\bf Contributions in system design:} To address the above challenges, we propose a new system design that leverages the advantages of image retrieval and 3D model-based localization.

\begin{itemize}
\item

In previous work \cite{majdik-iros-2013,zhang-3dpvt-2006,zamir-eccv-2010}, image retrieval has been applied for localization. The issue with these approaches are localization accuracy.  In particular, 
the location of the query is estimated through the geometric relationship between the queries and the retrieved images.
The accuracy depends on the performance of image retrieval. While some recent work has applied deep learning for image retrieval \cite{hoang:2017,do:2017,do:2016a,do:2016b}, applying them for resource-constrained mobile devices is challenging.
We have compared the accuracy of image retrieval based localization and
the results suggest that the accuracy could be inadequate (see Fig. \ref{overall-performance} in our experiments).

\item 3D model methods can achieve good localization accuracy \cite{li-eccv-2012,sattler-pami-2016}. However, these methods are not scalable: the memory requirement of storing the 3D point cloud of a large area is enormous. Furthermore, it is difficult to maintain a large 3D model:  updates in the city (e.g., newly constructed building) require substantial effort to  re-build a large 3D model   even with recent advances in Structure-from-Motion (SfM) \cite{snavely-siggraph-2006,wu-3dv-2013}.

\end{itemize}

Our proposed system design 
leverages the scalability advantage of image retrieval and accuracy of 3D model-based localization.
We propose to divide the region into sub-regions and construct 3D sub-models for the sub-regions. Sub-models are small and easier to be constructed,  and multiple sub-models can be constructed in parallel. Individual sub-models can be updated  without re-training other sub-models. 
Given a query image, in our proposed system, we apply image retrieval to identify the related sub-models. Then 2D-3D correspondence search is used for these sub-models.
Note that only the related sub-models need to be transferred into internal memory for processing, thus internal memory requirement is small.
Note that 
the work in \cite{arth-ismar-2009} also partitions  data/models into smaller parts. However, their work requires  GPS/WiFi or manual inputs to identify the relevant partitions.




{\bf Contributions in algorithms:} 
Furthermore, we make two main contributions in reducing the processing time and improving the accuracy of 2D-3D correspondence search.
First, 
we propose a  cascade hashing based search and re-ranking  using Product Quantization (PQ). 
Second, we propose a new one-many (1-M) RANSAC. The motivation of our 1-M RANSAC is as follows: Building facade usually has many repetitive elements (e.g., windows, balconies). These repetitive elements are similar in appearance, and the corresponding local descriptors are almost identical.
This complicates feature correspondence search. In particular, the correct correspondences may not be  in the top  rank, and they are mistakenly rejected when using conventional techniques
(See Fig. \ref{image_matching}
for some examples). This is an important issue for image-based localization. 
The goal of our proposed 1-M RANSAC is to reduce rejection of correct correspondences which are not in top rank, while requiring similar  computational complexity as conventional RANSAC.

Overall, through extensive experiments on workstations and mobile devices, we demonstrate that our proposed image-based localization system is faster, requires less memory, and is more accurate, comparing to other state-of-the-art.

In addition, we demonstrate our system on street view images of Google Street View (GSV) \cite{anguelov-joc-2010}.
GSV images can be potentially leveraged for practical applications that require extensive coverage of many cities in the world.
We investigate the potential of using GSV dataset for localization,  and this is important for practical localization systems. 
While there exists a number of prior works building their systems on GSV \cite{majdik-iros-2013,liu-acmm-2012,taneja-accv-2014,agarwal-iros-2015}, our work is different and focuses on camera pose estimation of images in a large-scale dataset using mobile devices. 
Note that GSV is a challenging dataset for pose estimation: common issues include low sampling rate, distortion, co-linear cameras, wide baseline,  obstacle objects (trees, vehicles) and query images taken using different devices at different  timing and conditions (distortion, illumination). Nevertheless, our results on a large GSV image dataset show that, via our proposed system design, new hashing-based cascade 2D-3D search and new one-many RANSAC, we can achieve a median error of less than 4m with average processing time less than 10s on a typical mobile device.

\section{Related Works}

\subsection{Image based Localization}

Early works of image-based localization can be divided into two main categories: retrieval based approach and 3D model-based approach (or direct search approach). Retrieval based methods \cite{zhang-3dpvt-2006,nister-cvpr-2006,zamir-eccv-2010,majdik-iros-2013,qian-tmm-2017} are closely related to image retrieval by matching query features against geo-tagged database images. This matching will result in a set of similar images according to the query. The query pose \cite{zhang-3dpvt-2006,majdik-iros-2013}, GPS (Global Positioning System) \cite{zamir-eccv-2010,li-tmm-2013} or POI (Places of Interest) \cite{qian-ieee-2017,qian-tip-2018} can be inferred from those references. This approach depends highly on the accuracy of image retrieval as it does not utilize the geometric information of 3D models. Unlike the retrieval based methods, the model-based approach directly performs the 2D-3D matching between the 2D features of the query image and the 3D points of the 3D model. A 3D model, which is a set of 3D points, is constructed from the given set of 2D images by using modern Structure-from-Motion (SfM) approaches \eg \cite{snavely-siggraph-2006}. This approach achieves more reliable results than the retrieval-based approach because it imposes stronger geometric constraints. Preferably, it holds more information about the 3D structure of the scene. Consequently, the camera pose can be computed from 2D-3D correspondences by RANSAC within Direct Linear Transform (DLT) algorithm \cite{hartley2003multiple} inside.

The representative works of 3D model based approach \cite{irschara-cvpr-2009,li-eccv-2010,sattler-iccv-2011,sattler-eccv-2012,li-eccv-2012}. \cite{irschara-cvpr-2009} use SfM models as the basis for localization. First, it performs image retrieval and then computes 2D-3D matches between 2D features in the query and 3D points visible in top retrieved images. Synthetic views of 3D points are generated to improve image registration. \cite{li-eccv-2010} compresses the 3D model and prioritizes its 3D points (given the prior knowledge from visibility graph) in 3D-2D correspondence search and this allows the "common" views to be localized quickly. \cite{sattler-iccv-2011} proposes the efficient prioritization scheme to stop the 2D-3D direct search early when it has detected enough number of correspondences. \cite{sattler-eccv-2012,li-eccv-2012} proposes two-directional searches from 2D image features to 3D points and vice versa, this approach can recover some matches lost due to the ratio test.

A recent trend in 3D model-based localization shifts the task of finding correct correspondences from the matching step to pose estimation step through leveraging of geometric cues. \cite{svarm-cvpr-2014} proposes an outliers filter with the assumption of a known direction of gravitational vector and the rough estimate of the ground plane in a 3D model. Consequently, the pose estimation problem can be cast into a 2D registration problem. Following the same setup like \cite{svarm-cvpr-2014}, \cite{zeisl-iccv-215} proposes a filtering strategy which is based on Hough voting in linear complexity. To reduce the computational time of the method, the authors exploit the verification step by using local feature geometry, such as the viewing-direction constraints or the scale and the orientation of 2D local features to reject false early matches before the voting. \cite{camposeco-cvpr-2017} proposes a two-point formulation to estimate the absolute camera position. This solver combines the triangulation constraint of viewing direction and toroidal constraints as the camera is known to lie on the surface of torus.

Rather than explicitly estimating the camera pose from 2D-3D matching, recent works have applied deep learning for this problem \cite{kendall-cvpr-2015,melekhov-iccvw-2017,kendall-cvpr-2017,walch-iccv-2017}. They directly learn to regress the camera pose (e.g. 6 Degree-of-Freedom (DOF)) from images. However, this approach may require further research to achieve the comparable accuracy of camera pose estimation as the 3D model-based approach. Besides, applying them for resource-constrained mobile devices is challenging.

\subsection{On-device systems}
 
All 3D model-based methods require a massive amount of memory to store SIFT descriptors. Due to memory constraints, loading a large 3D model on memory to perform the correspondence search is impractical. Some earlier works tried to build localization systems that run on mobile devices. \cite{arth-ismar-2009} keeps the 3D model out-of-core and manually divides it into multiple segments that fit into the memory capability of a mobile phone. However, this work is confined to small workspaces and requires the initial query image location with the support of WiFi, GPS,... or manual inputs need to be provided. The work is extended for outdoor localization \cite{arth-ismar-2011}, but prior knowledge of coarse location or relevant portions of pre-partitioned databases downloaded from wireless network is still needed.
\cite{ventura-tvcg-2014} and \cite{middelberg-eccv-2014} employ the client-server architectures. These methods first estimate the camera pose on devices, and further improve the pose estimation by aligning it with the global model to avoid the drift. While \cite{ventura-tvcg-2014} keeps part of the global model on device's memory to speed up the matching, \cite{middelberg-eccv-2014} reconstructs its own map of the scene and uses the global pose received from an external server to align to this map. \cite{lim-cvpr-2012} use Harris-corner detectors and extract two binary features for tracking and 2D-3D matching. It avoids excess computation via matching over a small batch of tracked keypoints only. \cite{lynen-rss-2015} implements a fast pose estimation and tracking entirely on a device. This work uses Inverted Multi-Index (IMI) \cite{babenko-cvpr-2012} for compressing and indexing 3D keypoints that allows storage of the 3D model into device memory. However, using this scheme may eliminate 3D points which are necessary to localize many difficult queries.

\subsection{Using Street View images for localization}

One of the difficulties in developing a large-scale image-based localization is data collection where ground-truth data, \eg camera pose or GPS, in real-world are required. Several on-device systems \cite{arth-ismar-2011, ventura-tvcg-2014, middelberg-eccv-2014, lim-cvpr-2012,lynen-rss-2015} have to collect their own dataset for experiments which are usually confined to small areas. Mining images from online photo collections like Flickr \cite{snavely-siggraph-2006} is an attractive solution. However, this undertaking is challenging due to noisy distortions distributed in the real world images. In addition, the coverage of images is often in popular places, \eg city landmarks. \cite{chen-cvpr-2011} approached by using cameras-mounted surveying vehicles to harness the street-level data in San Francisco. They published a dataset containing 150k high-resolution panoramic images of San Francisco to the community. \cite{majdik-iros-2013} uses GSV images to localize UAV by generating virtual views and matching images with strong viewpoint changes. \cite{taneja-accv-2014} performs tracking of vehicles inside the structure of the street-view graph by a Bayesian framework. This system requires compasses measurements and fixed cameras within many assumptions of video capturing conditions. \cite{agarwal-iros-2015} tracks the pose of a camera from a short stream of images and geo-register the camera by including GSV images into the local reconstruction of the image stream. Nearby panoramic images are determined by image retrieval with restrictions of locations inferred by GPS or cellular networks in the surrounding 1km area.

\section{Proposed System}


We first provide an overview of our proposed design for on-device large-scale localization system to overcome the constraints of memory and computation on a typical mobile device. Then, we discuss our main contribution of the 2D-3D correspondence search in speeding up the system.

\subsection{On-device localization system}

\begin{figure*}
	\centering
	\includegraphics[scale=0.85]{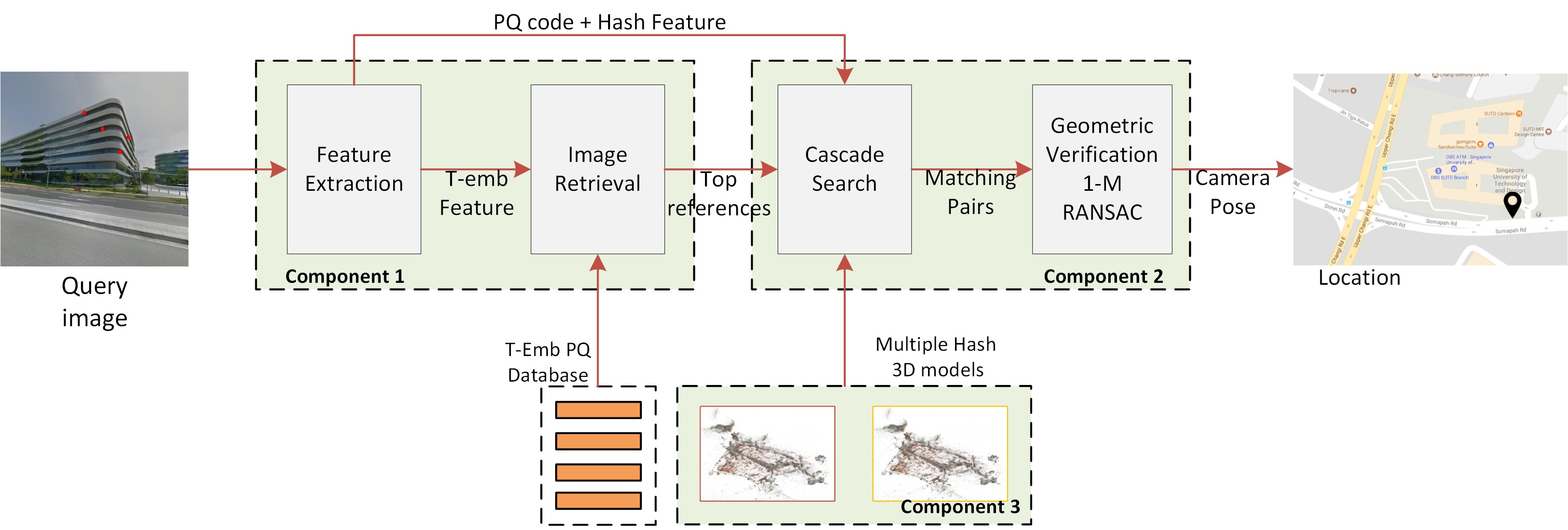}
	\caption{Overview of our proposed system with three main components. Image retrieval (IR) identifies reference images that are similar to the query image. The retrieved images indicate relevant 3D models. Then, camera pose is calculated by aligning the query image to these 3D models using cascade search and one-many (1-M) RANSAC.}
	\label{fig_system_overview}
\end{figure*}

We design our system in a hierarchical structure: we first divide the scene into smaller parts or \textit{segments}, then we index them using image-retrieval method to quickly find possible segments of the scene where the query image belongs, and finally, we localize the camera pose of the query by these selected segments using 3D model-based approach.
Our proposed system design aims to overcome the constraints of memory and computation (while preserving competitive accuracy) when using a large-scale dataset on a typical mobile device. We demonstrate our overall system via a large collection of GSV images for urban localization.
Our system has three main components (Fig. \ref{fig_system_overview}): (i) The first component is the set of 3D models to represent the scene. Instead of representing the entire 3D scene by a single model, we divide the scene into smaller segments and construct small 3D models from those segments. (ii) The second component uses image retrieval to identify similar images (or references) given a query as well as the 3D model candidates from those references. In this work, we apply the image-retrieval method proposed in \cite{jegou-cvpr-2014}, this method is memory-efficient, fast and accurate. (iii) The third component is the 2D-3D correspondence search and geometric verification. We propose a new cascade search and the one-many RANSAC to improve localization accuracy and reduce latency. These will be discussed in more detail. In this work,  we apply SIFT \cite{lowe-ijcv-2004} features as the input for both image retrieval and 2D-3D correspondence search, as SIFT has been demonstrated to be reliable and efficient in various applications: 3D reconstruction, image retrieval, and image-based localization. Note that other features can be used for our proposed pipeline.



\subsubsection{Scene representation using small 3D models}

\begin{figure*}
	\centering
	\includegraphics[scale=0.5]{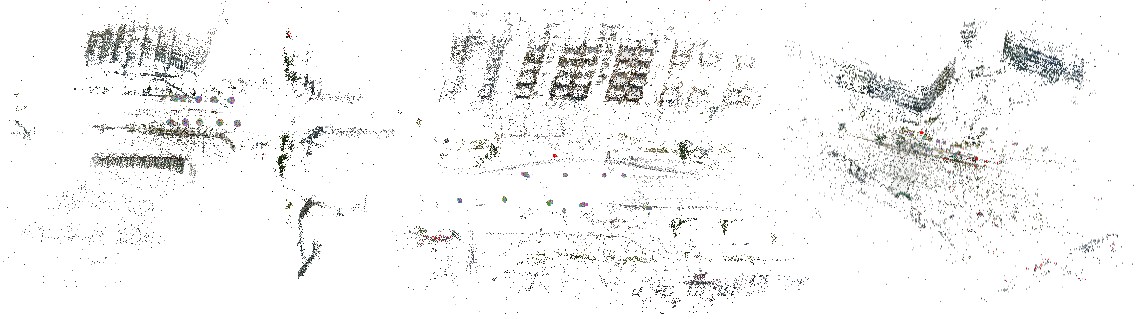}
	\caption{Examples of 3D models reconstructed by SfM. 
	}
	\label{fig_3d_model_examples}
\end{figure*}

\begin{figure}
	\centering
	\includegraphics[scale=0.3]{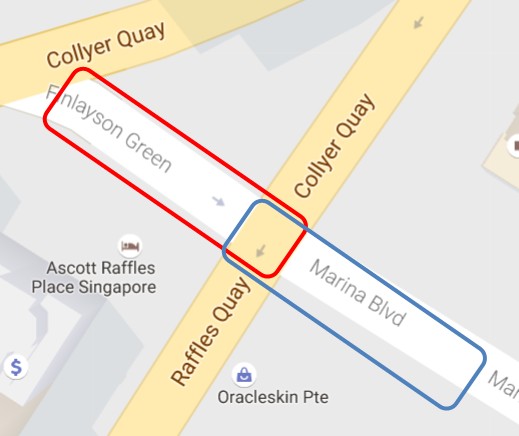}
	\includegraphics[scale=0.3]{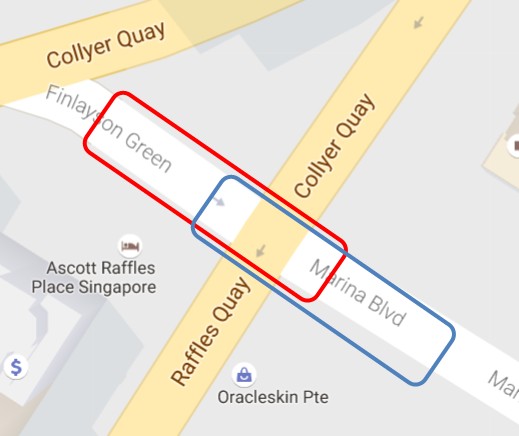}
	\caption{We represent the scene with overlapping segments, and build small 3D models for individual segments. We investigate the effect of overlapping on localization accuracy.}
	\label{fig:overlap-illustration}
\end{figure}

We demonstrate our overall system on a collection of Google Street View (GSV) \cite{anguelov-joc-2010} images. GSV is a very large image dataset. Constructing a single, large 3D model from such a large-scale dataset is computationally expensive. Moreover, it could be difficult to load such a large 3D model into the internal memory of mobile devices. In addition, representing the scene by a single model is inflexible: It is rather difficult to update a large model when some region of the city changes (\eg newly constructed buildings). Therefore, in our work, we divide the scene into smaller {\em segments} and build small 3D models for individual segments (Fig. \ref{fig:overlap-illustration}). Reconstruction of small 3D models can be performed in parallel, and this reduces the processing time needed build the scene models.  Moreover, provided that the corresponding small 3D models can be correctly identified, localization using small 3D models can achieve better accuracy as there exists less number of distracting 3D points. Furthermore, localization time can be reduced using small 3D models. We use 8-10 consecutive GSV placemarks to define a segment of the scene. As we sample 60 street view images per placemark, there are 480-600 images for a segment. These numbers are determined through experiments in Section \ref{gsv_model_design}. We use SIFT to detect keypoints for image datasets and Incremental SfM \cite{snavely-siggraph-2006,wu-3dv-2013} to reconstruct a 3D model from the images of a segment. See examples of our 3D models in Fig. \ref{fig_3d_model_examples}. Note that instead of the original SIFT descriptors of these 3D models, their hash code and quantized representation are stored. This reduces the memory requirement and speeds up the search. It will be discussed in Section \ref{2d_3d_matching}.

\subsubsection{Model indexing by image retrieval}

We also use the image retrieval (IR) in our framework. However, in contrast to the image retrieval based approach whose localization is sensitive to the resulted list, we use IR to identify the list of 3D models $\mathbb{M}_i$ for localizing the query image. In our case IR serves as the coarse search to limit the searching space for the second step (2D-3D correspondence search).

Let $\{I_j\}_{j=1:N}$ be the $N$ images in dataset. If the image $I_j$ was used to reconstruct 3D model $\mathbb{M}_i$, we set $r(\mathbb{M}_i,I_j) = 1$, otherwise $r(\mathbb{M}_i,I_j) = 0$. 
Given a query image $I_q$, image retrieval seeks top $N_t$ similar images from the dataset, namely $I_{j_1},I_{j_2},...,I_{j_{N_t}}$. $\mathbb{M}_i$ is a candidate model if $\exists I_{j_s}$: $r(\mathbb{M}_i,I_{j_s}) = 1$, $s=1:N_t$. 
Note that IR may identify multiple candidate models ($N_m$) for localizing the query image. 
In this case, the camera pose is estimated using the 3D model with the maximum number of 2D-3D correspondences (Section \ref{2d_3d_matching}). 
The summary of image retrieval is as follows: First, we extract SIFT features \cite{lowe-ijcv-2004} and embed them into high dimensional using Triangulation Embedding (T-embedding) \cite{jegou-cvpr-2014}. As a result, each image has a fixed-length T-embedding feature as a discriminative vector representation. We set the feature size to 4096.   To reduce the memory requirement and improve the search efficiency, we apply Product Quantization (PQ) with Inverted File (IVFADC)
\cite{jegou-pami-2011} to the T-embedding features. 
Details can be found in \cite{jegou-pami-2011,jegou-cvpr-2014}. Note that the PQ codes are compact.  As a result, we can fit the entire PQ codes of 227K reference images into the RAM of a mobile device.  Processing time for IR is less than 1s (GPU acceleration) for 227K reference images on a mobile device. Note that 227K images correspond to approximately 15km road distance coverage.

Using IR to index 3D models is memory efficient because only a few models are processed each time. On the other hand, performing 2D-3D correspondence search is more expensive due to matching between the query and $N_m$ models. This leads to our proposed idea of correspondence search which aims to reduce this computational complexity.

\subsection{Fast 2D-3D correspondence search}
\label{2d_3d_matching}

Our proposed method for 2D-3D correspondence search, namely Cascade Correspondence Search (CCS), consists of two parts: (i) an efficient 2D-3D matching that seeks top ranked list of nearest neighbors in cascade manner and (ii) a fast and effective RANSAC which helps to boost accuracy through exploitation of inliers from a large number of correspondences.

\begin{figure*}
	\centering
	\includegraphics[scale=0.58]{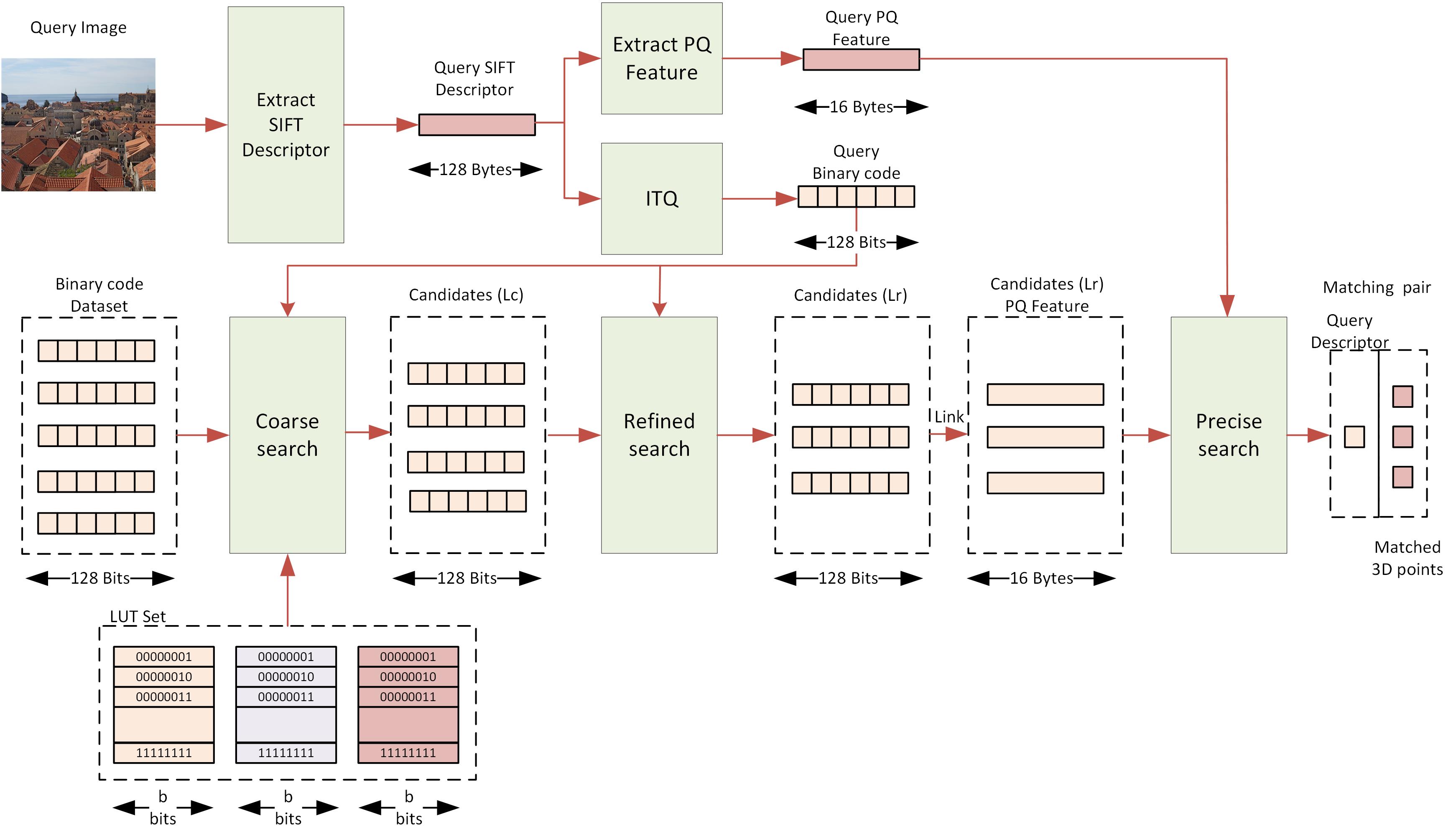}
	\caption{The pipeline of our cascade search. It consists of three main steps: coarse search (16-bit LUT), refined search (128-bit) and precise search (16-byte). SIFT descriptors (128 bytes) are compressed into 128-bit binary vectors. These vectors are used in the coarse search to quickly identify a short list of candidates. These candidates are then examined in the precise search with  PQ. Precise search  identifies  correspondences  for the next step, i.e., RANSAC.}
	\label{fig:fig_proposed_matching_scheme}
\end{figure*}

\subsubsection{Cascade search for 2D-3D matching}

Our method leverages the efficient computation of Hamming distance. We follow the Pigeonhole Principle on binary code \cite{norouzi-cvpr-2012} to further accelerate the search. The key idea is the following \cite{norouzi-cvpr-2012}: A binary code $\mathbf{h}$, comprising $d$ bits, is partitioned into $m$ disjoint sub-binary vectors,  $\mathbf{h}^{(1)},...,\mathbf{h}^{(m)}$ , each has $\left\lfloor \frac{d}{m}\right\rfloor$ bits. For convenience, we assume that $d$ is divisible by $m$. When two binary codes $\mathbf{h}$ and $\mathbf{g}$ differ at most $r$ bits, then, at least, one of $m$ sub-binary vectors, for example $\{\mathbf{h}^{(k)},\mathbf{g}^{(k)}\}, 1\leq k\leq m$, must differ at most $\left\lfloor \frac{r}{m}\right\rfloor$ bits. Formally, it can be written:

\begin{equation}
\left\Vert \mathbf{h}-\mathbf{g}\right\Vert _{H}\leq r \Rightarrow  \exists k \in [1,m]: \left\Vert \mathbf{h}^{(k)}-\mathbf{g}^{(k)}\right\Vert _{H}\leq\left\lfloor \frac{r}{m}\right\rfloor \label{pigeonhole}
\end{equation}

where $\left\Vert.\right\Vert_H$ is the Hamming distance.

The pipeline of our proposed 2D-3D matching method is shown in Fig. \ref{fig:fig_proposed_matching_scheme}. The method includes three main steps: coarse search, refined search, and precise search. Two first steps are to quickly filter out a shorter list of candidates from $N_p$ 3D points' descriptors, the last step to precisely determine the top-ranked list. Let $d=128$ be the feature dimension of SIFT descriptors. Given a 3D model and its points' descriptors, each descriptor $\mathbf{d} \in \mathbb{R}^{d \times 1}$ is pre-mapped into binary vector $\mathbf{h}$ in Hamming space $\mathbb{B}^{d \times 1}$: $\mathbf{h} = \mathrm{sign}(\mathbf{W}\mathbf{d})$, where $\mathbf{W}$ is the transformation matrix, which can be learned via objective minimization:

\begin{equation}
\arg\min_{\mathbf{W},\mathbf{H}} \left\Vert \mathbf{H} - \mathbf{W}\mathbf{D}\right\Vert_F^2
\end{equation}

where $\left\Vert.\right\Vert_F$ is Frobenius norm. $\mathbf{D}, \mathbf{H}$ are matrices of all point descriptors of 3D model (one descriptor per matrix's column) and its binarized code after transformation respectively. We solve the optimization problem by ITQ \cite{gong-cvpr-2011}. Given the learned hash function, all descriptors of the model are mapped into binary vectors and we store those vectors instead of the original SIFT descriptors.

{\bf Coarse search:} We follow the principle (\ref{pigeonhole}) to create a LUT (Lookup Table) based data structure for fast search. We split binary vector $\mathbf{h}$ into $m$ sub-vectors $\{\mathbf{h}^{(k)}\},k\in[1:m]$ of $b$ bits ($m*b=d$). In our work, we only select candidates differ at most $r = m-1$ bits from the query ($\left\lfloor \frac{r}{m}\right\rfloor = 0$). In other words, a candidate's binary vector is potentially matched to the query's iff at least one of their sub-vectors are exactly the same. For training, we create $m$ LUTs, where $\mathbf{LUT}^{(k)}$ for the sub-vector $\mathbf{h}^{(k)}$, and each LUT comprises of $K_b = 2^b$ buckets. One bucket links to a point-id list of 3D points that are assigned to buckets according to their binary sub-vectors. For searching, a query descriptor is first mapped into Hamming space and was divided into $m$ sub-binary vectors as above. And then looking up into the $\mathbf{LUT}^{(k)}$ to find a certain bucket that matches the binary code of $\mathbf{h}^{(k)}$. This results in the point-id list $\mathbf{L}^{(k)}$:

\begin{equation}
\mathbf{L}^{(k)} = \mathbf{LUT}^{(k)}(\mathbf{h}^{(k)}), k=1,..,m
\end{equation}

\begin{table}
\footnotesize
\centering
\caption{The trade-off between the number of candidates $\mathbf{L}_C$ and the size of LUT experimented on Dubrovnik dataset.}
\begin{tabular}{|c|c|c|}
\hline
\textbf{Method($m$,$b$)} & $\mathbf{L}_C$ & \textbf{LUT size} \tabularnewline
\hline 
\hline 
ITQ(4,32) & - & $4 \times 2^{32} \times 4$ = 64GB \tabularnewline
\hline 
\textbf{ITQ(8,16)} & $\sim$ \textbf{5K} & \textbf{$8 \times 2^{16} \times 2$ = 1048KB} \tabularnewline
\hline 
ITQ(16,8) & $\sim 60K$ & $16 \times 2^{8}$ = 4096B\tabularnewline
\hline 
LSH(8,16) & $\sim 20K$ & $8 \times 2^{16} \times 2$ = 1048KB \tabularnewline
\hline 
\end{tabular}
\label{tbl-hashing-table-storage}
\end{table}

Next, merging $m$ point-id list to have the final list of coarse search $\mathbf{L}_C = [\mathbf{L}^{(1)},...,\mathbf{L}^{(m)}]$. 
By using LUT, the search complexity of $\mathbf{h}^{(k)}$ is constant $O(1)$ when retrieving the point-id list $\mathbf{L}^{(k)}$. This step results in a short list that $\mathbf{L}_C$ contains $|\mathbf{L}_C|$ candidates for the next search. It is important to choose appropriate values of $m$ and $b$ for the trade-off between the memory requirement of LUT and computation time (which depends on the length of $\mathbf{L}_C$ that requires Hamming distance refining). As shown in Table \ref{tbl-hashing-table-storage}, we map descriptors to binary codes by using ITQ with different settings and also replace it with LSH \cite{charikar-acm-2002} based scheme \cite{cheng-cvpr-2014}. ITQ($m=4$,$b=32$) is impractical due to over-large size requirement of LUTs.  ITQ($m=16$,$b=8$) results in too many candidates, which slows down the refined search. ITQ($m=8$,$b=16$) is the best option, results in the short list, and requires a small amount of LUT memory (excluding the overhead memory of descriptors indexing). Using multiple lookup table using LSH \cite{cheng-cvpr-2014} results in the longer list ($\sim4\times$ of ITQ) of candidates, which means that learning the hash mapping from data points by ITQ is more efficient than a random method LSH in our context. This is consistent with our experiments conducted later in Fig. \ref{dubrovnik-pq-methods}.

{\bf Refined search:} In this step, we use full $d$-bit code $\mathbf{h}$ to refine $\mathbf{L}_C$ list to pick out a shorter list $\mathbf{L}_R$ ($\mathbf{L}_R \leq 50$). First, we compute exhaustively the Hamming distance between the $d$-bit code of query to that of $\mathbf{L}_C$ candidates. Then, candidates are re-ranked according to these distances. Computing Hamming distance is efficient because we can leverage low-level machine instructions (XOR, POPCNT). Computing Hamming distance of two 128-bit vectors is significantly faster ($\geq$ 30$\times$) than the Euclidean distance of SIFT vectors and accelerates ($\geq$ 4$\times$) ADC (Asymmetric Distance Computation) \cite{jegou-pami-2011} on our machine. Furthermore, Hamming distance of $d$-bit code has the limited range of [0, 128], which allows us to build the online LUT during the refined search. As such, selecting top candidates $\mathbf{L}_R$ search is accelerated. However, the limited range prevents us to precisely rank candidates. That leads to the last step of our pipeline.

{\bf Precise search:} The purpose of the precise search is to get $\mathbf{L}_R$ ranked better so that we can choose the best candidate or remove outliers of matches before applying geometric verification. Furthermore, we can consider their order as an useful prior information. It plays an important role to reduce the complexity of pose estimation (discussed in Section of Geometric Verification). The approximated Euclidean distance by ADC of PQ \cite{jegou-pami-2011} is used. The match between a query feature and a 3D point is established if the distance ratio from the query to the first and second candidates passes the ratio test $\nu_h$ \cite{lowe-ijcv-2004}; otherwise, they are rejected as outliers. The sub-quantizers of PQ are trained once from an independent dataset, SIFT1M \cite{lowe-ijcv-2004}, and used in all experiments. In this step, we need to store PQ codes in addition to hashing code of two previous steps.

In addition to \cite{norouzi-cvpr-2012}, some form of cascade hashing search has been applied for image matching \cite{cheng-cvpr-2014}. In this work, we apply it for 2D-3D matching and propose several improvements beyond the work of  \cite{norouzi-cvpr-2012,cheng-cvpr-2014}:

\begin{itemize}
\item
In our work, since the 3D models are built off-line and SIFT descriptors for 3D points are available during off-line processing, we propose to train an unsupervised data-dependent hash function to improve matching accuracy. \cite{norouzi-cvpr-2012,cheng-cvpr-2014} make use of  Locality Sensitive Hashing (LSH) \cite{charikar-acm-2002}, which has no prior assumption about the data distribution. In contrast, we apply Iterative Quantization (ITQ) \cite{gong-cvpr-2011}, in which the hash function is learned from data.
\item
We use a single hash function of ITQ for mapping from 128 bytes SIFT to $d$ bits binary vector. Splitting the long $d$ bits code into $m$ short-codes of $b$ bits to construct $m$ lookup tables (LUT) for coarse search and use full $d$ bit vector for the refined search. In contrast, \cite{cheng-cvpr-2014} created multiple lookup tables using LSH with short-codes. These tables are independent and built from random projection matrices that return the long list of candidates, hence slowing down the next step of refined search (discussed later in Table \ref{tbl-hashing-table-storage}).
\item
We add the precise search layer to the hashing scheme and propose to use Product Quantization (PQ) \cite{jegou-pami-2011}, a fast and memory efficient method for precise search. Consequently, our work combines hashing and PQ in a single pipeline to leverage their strengths: Binary hash code enables fast indexing via Hamming distance-based comparison, while PQ achieves better matching accuracy. They are both compressed descriptors. Without this  precise search step,  accuracy is significantly reduced. However, using the original SIFT descriptor for this step \cite{cheng-cvpr-2014} requires considerable amount of  memory storage (128-byte to store a SIFT descriptor). As will be discussed, using PQ in our method can achieve similar accuracy, but our method requires only 16 bytes per descriptor. This reduces memory requirement by about 8 times as compared to the original SIFT. In our experiments, we compare our search method to the method in \cite{cheng-cvpr-2014}, and we use  PQ for the last step in both methods for fair comparison.

\end{itemize}

\subsubsection{Prioritization and pose estimation}

In addition to above improvements from cascade hashing search, we propose a prioritizing scheme and the fast one-many RANSAC to significantly speed up the search, while preserving competitive accuracy.

{\bf Prioritization: } 
Finding all matches between 2D features and 3D points to infer camera pose is expensive because the query image can contain thousands of features. In practice, the method can stop early once found a sufficient number of matches \cite{sattler-iccv-2011}. Therefore, we perform a prioritized search on descriptors of the 2D image as follows: given a query descriptor, the coarse search returned the point-id list $\mathbf{L}_C$. We first continue the refined and precise search with those query features having shorter list $|\mathbf{L}_C|$. A correspondence is established if the nearest candidate passes the ratio test with threshold $\nu_h$ on precise search. We stop the search once $N_{early}=100$ correspondences have been found.  This is an important proposed technique: in our context, it is not necessary to find all 2D-3D correspondences for localization. It is sufficient for localization as long as a certain number of correspondences are found. Results show that this scheme can significantly accelerate the system  (about $\sim10\times$) and incur minimal accuracy degradation. The evaluation is demonstrated on the Dubrovnik dataset in Table \ref{dubrovnik-state-of-the-art}.


{\bf Pose estimation by one-many RANSAC: }
\label{geometric_verification}

\begin{figure}
	\centering
	\includegraphics[scale=0.205]{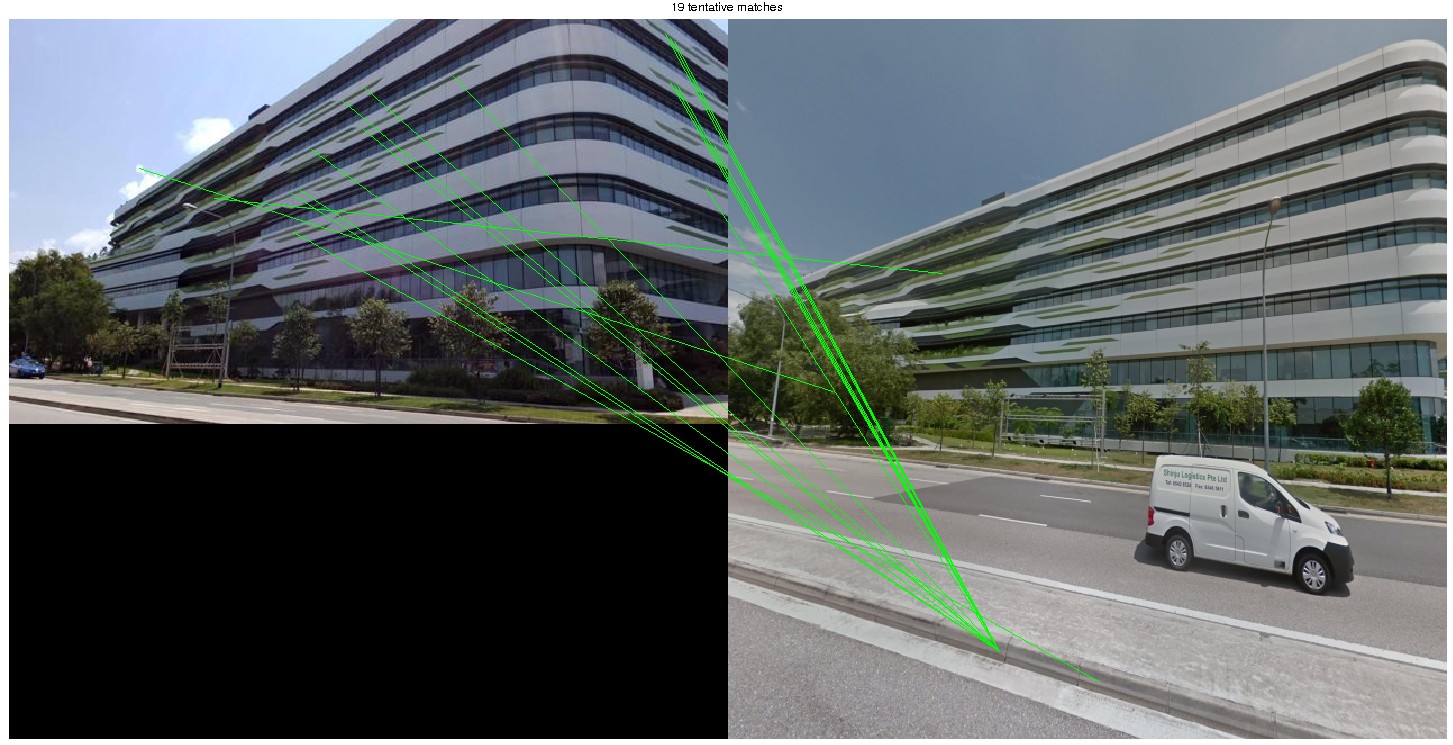}
	\includegraphics[scale=0.2625]{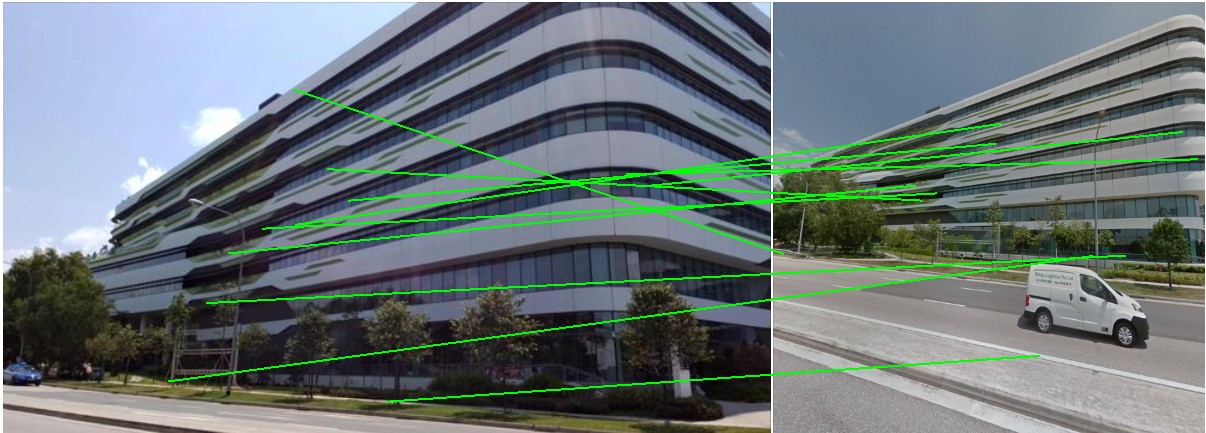}	
	\includegraphics[scale=0.205]{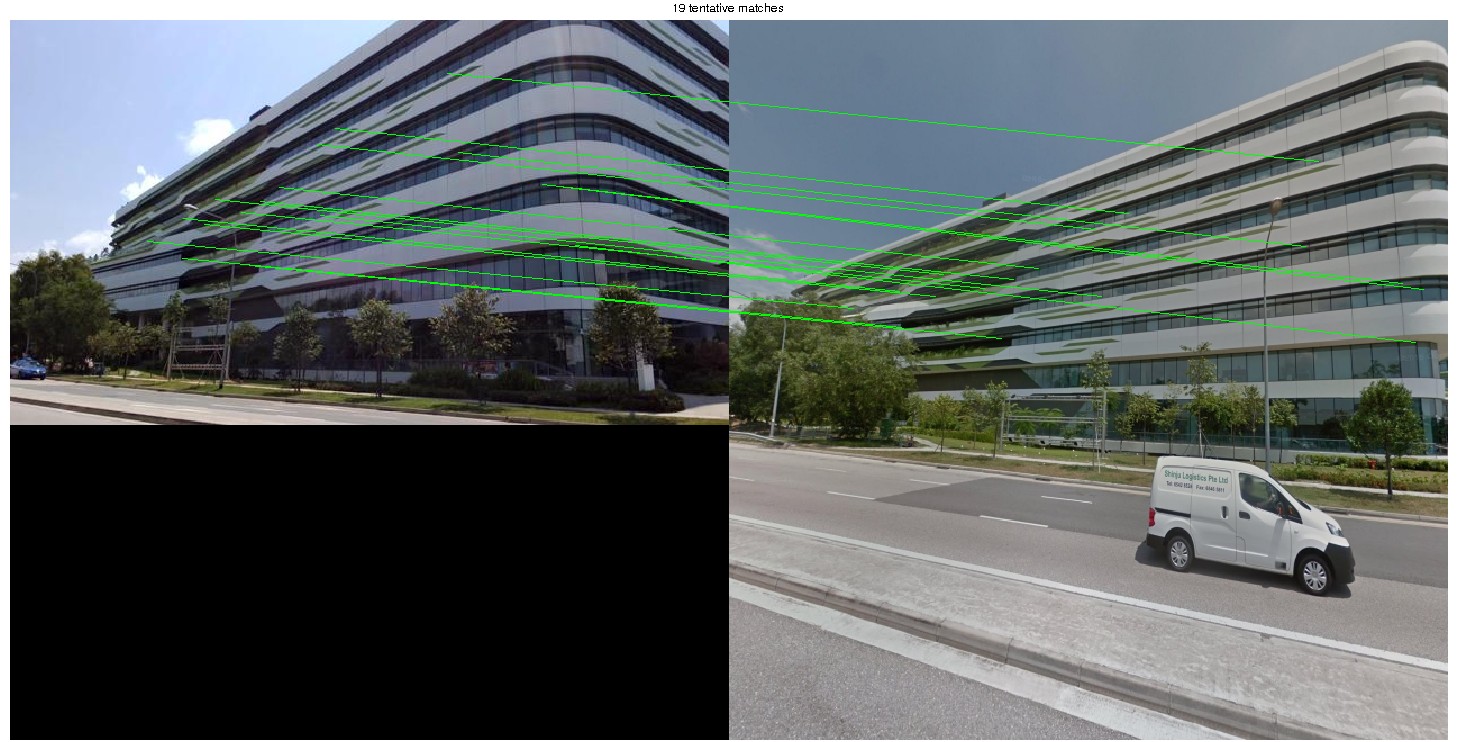}
	\caption{Conventional RANSAC, RANSAC of VisualSFM, and our 1-M RANSAC for image matching.}
	\label{image_matching}
\end{figure}

One of the long-standing problems in correspondence matching is the  problem of rejecting correct matches using the ratio test.
The problem is more severe in image-based localization:  Building facade usually has many repetitive elements (e.g., windows, balconies). These repetitive elements are similar in appearance, and the corresponding local descriptors are almost identical. 
Please refer to Fig. \ref{image_matching} for some examples. In our work, we propose to retain more potential matching candidates as a feature in the image may have multiple matching candidates in the 3D model. We propose to use the geometric constraints to filter out the outliers. However, this poses problem to conventional RANSAC as it is very computationally expensive to iterate on many pairs of candidates. In particular, we need to perform this on resource-constrained mobile devices. Our proposed one-many (1-M) RANSAC is a new solution to this problem. We use the hypothesis set to create the hypothesis model, and use the verification set to validate the model. In addition, we use the pre-verification step to quickly reject bad hypothesis models. 
Note you will see in the later results that on average our 1-M RANSAC can increase the number of correspondences by a factor of two or more. The details of our proposed algorithm are as follows.

After 2D-3D matching, traditionally, one query descriptor has a maximum of one 3D point correspondence. Those correspondences (one-one matches) are then filtered out by geometric constraints, \eg RANSAC within 6-DLT algorithm inside. Empirically, we made two observations: (i) ratio test $\nu_h$ tends to reject many good matches (ii) good candidates are not always highest-ranked in the list $\mathbf{L}_R$. It is probably due to repetitive features in buildings and it is a common issue of localization in urban environment \cite{sattler-pami-2016,zeisl-iccv-215}. Therefore, relaxing the threshold to accept more matches (one-many matches), and filtering wrong matches by using geometric verification seem to be potential solutions. Recent works \cite{svarm-cvpr-2014,zeisl-iccv-215} use these approaches, but their geometric solvers are too slow for practical applications.

To address this issue, we propose a fast and effective one-many RANSAC as follows: First, we relax the threshold $\nu > \nu_h$ to accept more matches and keep one-many candidates per query descriptor. We compute one-many matchings: given one query feature, we accept $M$ top candidates in $\mathbf{L}_R$ list where $d_0 / d_1 < \nu$. $d_0$ and $d_1$ are the first and second smallest distances of the query to $\mathbf{L}_R$ candidates. However, processing all these matchings leads to an exponential increase in the computational time of RANSAC due to the low rate of inliers.

We avoid this issue by considering its subset to generate the hypothesis. Consequently, we propose two different sets of matchings in the hypotheses and verification stages of RANSAC. The first set contains the one-one (1-1) matchings that pass the ratio test with threshold $\nu_h$. The second set of matchings contains one-many (1-$M$) matchings found by the relaxed threshold as mentioned above. We propose to use the first set to generate hypotheses and the second set for verification. We found that using relaxed threshold and 1-$M$ matchings in verification can increase the number of inliers, leading to the accuracy improvement. We speed up our method by applying the pre-verification step like \cite{chum-pami-2008}, which based on Wald's theorem of sequential probability ratio test (SPRT). This step is helpful to quickly reject the "bad" samples before the full verification.

The details are as such: let assume that 2D-3D matching has found matches between 2D keypoint queries $\mathbf{Q} = \{\mathbf{q}_i\}$, and 3D points of the model $\mathbf{P} = \{\mathbf{p}_{ij}\}$, $i=1:N_q$, where $N_q$ is the number of 2D queries. $\mathbf{q}_i$ is the 2D coordinate of $i$-th query and $\mathbf{p}_{ij}$ is the 3D coordinate of $j$-th matches of $i$-th query. Those matchings (verification set) found by 2D-3D matching with the relaxed threshold $\nu$: $d_{i0}/d_{ij} < \nu$, where $d_{i0}$ is the first and $j$-th distances of the candidate list to the query $\mathbf{q}_i$: $\mathbf{P}_i = \{\mathbf{p}_{ij}\},j=1:|\mathbf{P}_i|$. $|\mathbf{P}_i|$ is the number of candidates matched to each query $\mathbf{q}_i$, and $|\mathbf{P}_i| \leq M$. Without the loss of generality, $\mathbf{p}_{ij}$ were sorted in ascendant of ADC distances from 2D-3D matching. Our hypothesis set is a subset of verification set. It has $N_h (\leq N_q)$ 1-1 matchings $\{\mathbf{q}_{i_k},\mathbf{p}_{{i_k}{1}}\}$, $i_k \in \left[1:N_q\right]$, $k=1:N_h$ which passed the strict threshold $\nu_h$: $d_{i_k0}/d_{i_k1} < \nu_h$.

In our algorithm, $\epsilon$ indicates the probability $p(1|H_g)$ that a random match is consistent with a ``good" model. This probability is initialized: $\epsilon=\frac{s}{N_h}$. $\delta$ indicates the probability $p(1|H_b)$ of a match being consistent with a "bad" model. This probability is initialized with a small value: $\delta=0.01$. The probability of rejecting a "good" sample ($\alpha = 1/A$), where $A$ is the decision threshold (discussed later). Here, $H_g$: the hypothesis that the model is "good", and $H_b$: the alternative hypothesis that the model is "bad".

The details of proposed algorithm are presented in Fig. \ref{alg-1m-ransac} and in three main steps as follows: 

First, in the hypothesis step, a model sample ($s$ correspondences) is randomized from the hypothesis set $\{\mathbf{q}_{i_k},\mathbf{p}_{{i_k}{1}}\}$, $i_k \in \left[1:N_q\right]$, $k=1:N_h$. $s = 6$ indicates the minimum number of correspondences that can be used to estimate the model parameters $\theta$ using 6-DLT algorithm. $\theta$ is a 3D-2D projection matrix, which can map 3D coordinates to 2D keypoints on the image plane. The model parameters $\theta$ is computed from $s$ random correspondences $(\mathbf{q}_{i_k},\mathbf{p}_{i_k1})$. This model will be validated whether it is a ``good" or ``bad" model in the pre-verification step. Randomizing samples from the hypothesis set, which is much smaller than the verification set, allows our RANSAC running fast enough.

Second, the pre-verification step further improves the processing speed because this step can quickly validate whether the model is ``good" or ``bad" after a small number of iterations. Hence, if the model is considered to be a ``bad" model, it will be better off re-generating new samples to avoid consuming time than continue the testing. In this step, we use correspondences from the hypothesis set for the pre-verification, $\rho_1$ is to check whether one correspondence is consistent with the estimated parameters of model $\theta$. The correspondence is consistent with the model, when the Euclidean distance between the query $\mathbf{q}_{i}$ and its 2D projection of $\mathbf{p}_{{i}{1}}$ is smaller than a threshold (eg. 4 pixels in the published code of ACS (Active Correspondence Search) \cite{sattler-eccv-2012}). We formulate this operator by $\rho_1$. The model is pre-verified via the likelihood ratio $\lambda_k$ computed from two conditional probabilities. If $\rho_1 = 0$ (the observation is not fitted/consistent to the model), the likelihood is updated with the ratio $\frac{1-\delta}{1-\epsilon}$ from previous iteration. Otherwise, it is updated with the ratio $\frac{\delta}{\epsilon}$. If $\lambda_k$ is higher than the decision threshold $A$, the model is likely to be "bad" and the pre-verification stops. In contrast, if the model is likely to be ``good", testing is continued. When the model is ``bad", some parameters $\delta$ and $A$ may be re-computed and a new sample in the hypothesis step is re-generated.

Third, if the model is likely to be ``good", all correspondences are checked with this model to locate the inliers. This verification step projects the correspondences $\mathbf{P}_i$ into the 2D image plane, and measures their Eulidean distances to the query $\mathbf{q}_i$. The correspondence $\mathbf{p}_{ij}$ passes the test if the Euclidean distance of its projection and $\mathbf{q}_{i}$ is smaller than the threshold. We formulate the verification as follows: if there exists $\mathbf{p}_{ij} \in \mathbf{P}_i$ passes the test, $\rho_{M}(\theta,\{\mathbf{q}_{i},\mathbf{P}_i\})$ is set 1, otherwise 0. In other words, $\rho_{M}(\theta,\{\mathbf{q}_{i},\mathbf{p}_{{i}{j}}\})=1$ if $\exists k$, $\rho_{1}(\theta,\{\mathbf{q}_{i},\mathbf{p}_{{i}{k}}\})=1$. The total cost $C \gets \sum_i \rho_{M}(\theta,\{\mathbf{q}_{i},\mathbf{P}_i\})$ is used to decide whether a new model is accepted or ignored. Validating tentative matches $\mathbf{P}_i$ of the query $\mathbf{q}_{i}$ is important in our RANSAC because the lower-ranked matches of $\mathbf{P}_i$ still have chances to be potentially chosen as a good correspondence. It is a minor change but improves the accuracy substantially.

Here, $C$ and $\theta$ are the cost (or the number of inliers) and model parameters respectively. If this cost $C$ is better than the optimal cost $C*$ (minimum cost obtained from previous iterations), it is a good model. As such, $C*$ and $\theta$ are updated and the probability $\epsilon$, the decision threshold $A$ and the number of iterations $\mu$ are re-computed. The adaptive decision threshold $A = A(\delta,\epsilon)$ is computed from probabilities $\delta$ and $\epsilon$ similar to \cite{chum-pami-2008}. $A$ is the decision threshold to make one out of three decisions for each observation: reject a ``bad" model, accept a ``good" model, or continue testing. This threshold is estimated using the SPRT theorem \cite{wald-tams-1945}. $\mu$ is the number of tested samples before a good "model" is drawn and not rejected. $\mu$ is computed from geometric distribution: $\mu=\frac{1}{\epsilon^s*(1-\alpha)}=\frac{1}{\epsilon^s*(1-\frac{1}{A})}$. It indicates that we need more iterations ($\mu$ is large) for testing if the probability of accepting a ``good" model is low ($\epsilon^s$ is small) and/or the probability of rejecting a ``good" is high ($\alpha=1/A$ is high), and vice versa.

\begin{figure}[!t]
\begin{algorithmic}[1]



\footnotesize
\Procedure{One-Many-RANSAC}{$\mathbf{Q}$, $\mathbf{P}$, $\{i_k\}_{k=1}^{N_{h}}$}

\State $\epsilon \gets p(1|H_g) = \frac{s}{N_h}$
\State $\delta \gets p(1|H_b) = 0.01$
\State $A \gets A(\delta, \epsilon)$
\State $\mu \gets \frac{1}{\epsilon^s*(1-\frac{1}{A})}$
\State $nr \gets 0$ \Comment{the number of rejected times}
\State $iter \gets 0$ \Comment{the number of iterations}
\While{$iter \leq \mu$}
\State $iter \gets iter + 1$
\State{\textbf{I. Hypothesis}}
	\State{Select a random sample of minimum size  $s$ from hypothesis set $\{\mathbf{q}_{i_k},\mathbf{p}_{{i_k}{1}}\}$, $i_k \in \left[1:N_q\right]$, $k=1:N_h$.}
	\State{Estimate model parameters $\theta$ fitting the sample.}
\State{\textbf{II. Pre-verification}}
    \State $k=1$
    \State $\lambda_0 = 1$
	\While{$k<=N_h$}
	    \State Let $\rho_{1} \gets \rho_{1}$($\theta$,$\{\mathbf{q}_{i},\mathbf{p}_{{i}{1}}\}$) \Comment{$0$ or $1$}        
        \State $\lambda_k \gets \lambda_{k-1} * (\rho_{1} * \frac{\delta}{\epsilon} + (1-\rho_{1})*\frac{1-\delta}{1-\epsilon})$

        \If{$\lambda_{k} > A$}
        	\State \texttt{bad\_model = true} \Comment{Reject sample}
        	\State \texttt{break}
        \Else
        	\State $k \gets k+1$
        \EndIf        
	\EndWhile
	\If{\texttt{bad\_model}}
	    \State $nr \gets nr + 1$ 
		\State $\hat{\delta} \gets \delta * \frac{nr-1}{nr} + \epsilon * nr$ \Comment{Re-estimate $\delta$}
		\If {$|\delta - \hat{\delta}| > 0.05$}
			\State $\delta \gets \hat{\delta}$ \Comment{Update $\delta$}
			\State $A \gets A(\delta, \epsilon)$ \Comment{Update $A$}
		\EndIf
		\State \texttt{continue}
	\EndIf
\State{\textbf{III. Verification}}
	\State Compute cost $C \gets \sum_i \rho_{M}(\theta,\{\mathbf{q}_{i},\mathbf{P}_i\})$
	\If{$C^* \leq C$}
		\State $C^* \gets C$, $\theta^* \gets \theta$ \Comment{Update good model}
		\State $\epsilon \gets C^**\frac{l_R}{N_q}$ \Comment{Update $\epsilon$}
		\State $A \gets A(\delta, \epsilon)$ \Comment{Update $A$}
		\State $\mu \gets \frac{1}{\epsilon^s*(1-\frac{1}{A})}$ \Comment{Update $\mu$}
	\EndIf
\EndWhile
\EndProcedure
\end{algorithmic}
\caption{The algorithm of our proposed RANSAC.}
\label{alg-1m-ransac}
\end{figure}

In addition to 2D-3D matching, our idea can also be used for conventional image matching. For example, Fig. \ref{image_matching} shows that 
with a building of many repetitive features, the classical RANSAC and VisualSFM's RANSAC fail, while our 1-M RANSAC still works in this case.

\normalsize

\section{Experimental Results}
\label{experiment_results}

We conduct experiments to validate our CCS method and the overall system. Specifically, we adopt four benchmark datasets: Dubrovnik \cite{li-eccv-2010}, Rome \cite{li-eccv-2010}, Vienna \cite{irschara-cvpr-2009}, and Aachen \cite{sattler-bmvc-2012}, to evaluate our correspondence search method and compare it againist the state-of-the-art. These four datasets are commonly used in earlier works \cite{li-eccv-2010,sattler-iccv-2011,li-eccv-2012} for evaluating the robustness of 2D-3D matching or 2D-3D correspondence search. Aachen images are collected different times and seasons day-by-day in the two-year period because it is important to evaluate the robustness of method against different times or seasons. We use these datasets to compare our correspondence search method to the state-of-the-art. Table \ref{standard-dataset-info} provides some information about these datasets. Then, we validate our on-device system design with the image collection of GSV. Our GSV dataset has 227K training images and 576 mobile queries. It is used to evaluate our image retrieval approach, and also the entire system (image retrieval and correspondence search (or localization)). 

Experiments are conducted on our workstation: Intel Xeon Octa-core CPU E5-1260 3.70GHz, 64GB RAM, and Nvidia Tablet Shield K1. 
We use ``mean descriptors" for each 3D point in all experiments.  We have three different settings for our method: Setting 1 uses traditional RANSAC (CCS), Setting 2 uses new 1-M RANSAC scheme (CCS + R$_{1-M}$) and Setting 3 indicates our method with the new RANSAC and prioritizing scheme included (CCS + P + R$_{1-M}$). Here: CCS, P and R$_{1-M}$ stand for Cascade Correspondence Search, Prioritizing, and One-Many RANSAC respectively. In the next sections, we will first evaluate our 2D-3D matching method and compare it to earlier works on benchmark datasets. Subsequently, we validate our system design on the GSV dataset.

\begin{table}
\small
\centering
\caption{Standard datasets for the evaluation of 2D-3D correspondences search.}
\begin{tabular}{|c|c|c|c|c|c|c|}
\hline
\textbf{Dataset} & \textbf{\#Cameras} & \textbf{\#3D Points} & \textbf{\#Descriptors} &  \textbf{\#Queries}\tabularnewline
\hline
\hline
Dubrovnik & 6044 & 1,886,884 & 9,606,317 & 800 \tabularnewline
\hline
Rome & 15,179 & 4,067,119 & 21,515,110 & 1000 \tabularnewline
\hline
Aachen & 3047 & 1,540,786 & 7,281,501 & 369 \tabularnewline
\hline
Vienna & 1324 & 1,123,028 & 4,854,056 & 266 \tabularnewline
\hline
\end{tabular}
\label{standard-dataset-info}
\end{table}

\subsection{Hashing based 2D-3D matching}

In this section, we evaluate our cascade search method and compare it against the other search methods. We then show the computational improvements (while remaining the competitive accuracy) when it is combined with our prioritizing technique and new proposed RANSAC algorithm.

\subsubsection{Hashing-based Matching}
\label{2d_3d_matching_experiments}

\begin{figure}[t]
	\centering
	\includegraphics[scale=0.4]{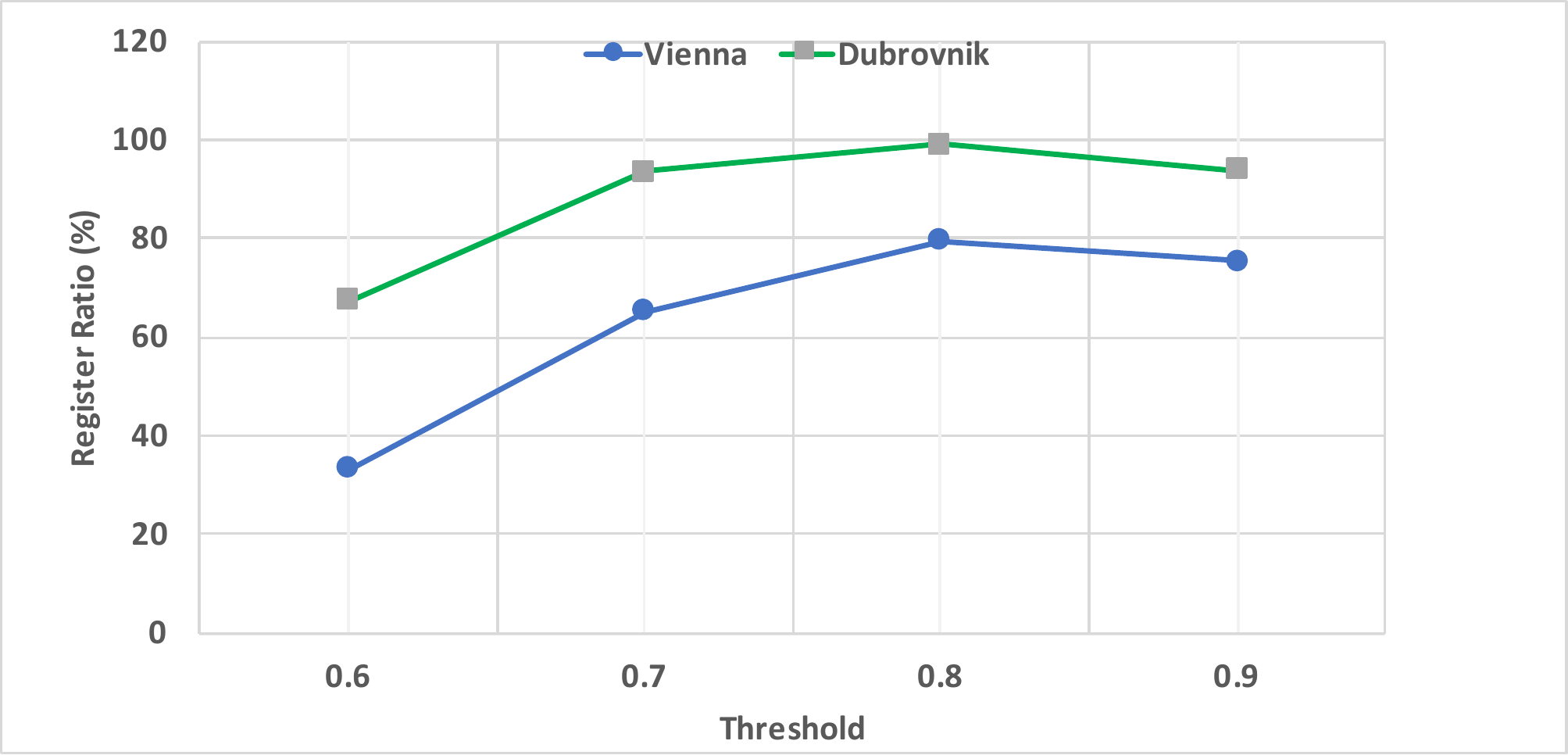}
	\caption{Studying the influence of test ratio thresholds on Dubrovnik and Vienna datasets. This experiment determines the good ratio threshold for precise search. Results show that threshold $\nu_h = 0.8$ achieves the highest registration rate. In the figure, the horizontal axis indicates the value of thresholds, and the vertical axis indicates the percentage of registered images.}
	\label{ratio_test_thresholds}
\end{figure}

The first experiment is used to determine a good test ratio threshold for precise search. It is conducted on the Dubrovnik and Vienna datasets. We use ADC with Inverted File \cite{jegou-pami-2011} and the number of coarse quantizer $K_c = 256$, $16$ sub-vectors of SIFT, the number of sub-quantizers $K_{pq} =2^8$, and the number of neighboring cells visited $w=8$. We use small $K_{c}$ and large $w$ to ensure that quantization does not significantly affect the overall performance. In this experiment, we fix 5000 iterations to attain the same probable result in multiple runs with RANSAC. A query image is ``registered" if at least twelve inliers found, same as \cite{li-eccv-2010}. This experiment suggests the threshold $\nu_h = 0.8$ is a good option, Fig. \ref{ratio_test_thresholds}.

\begin{figure}[t!]
	\centering
	\includegraphics[scale=0.46]{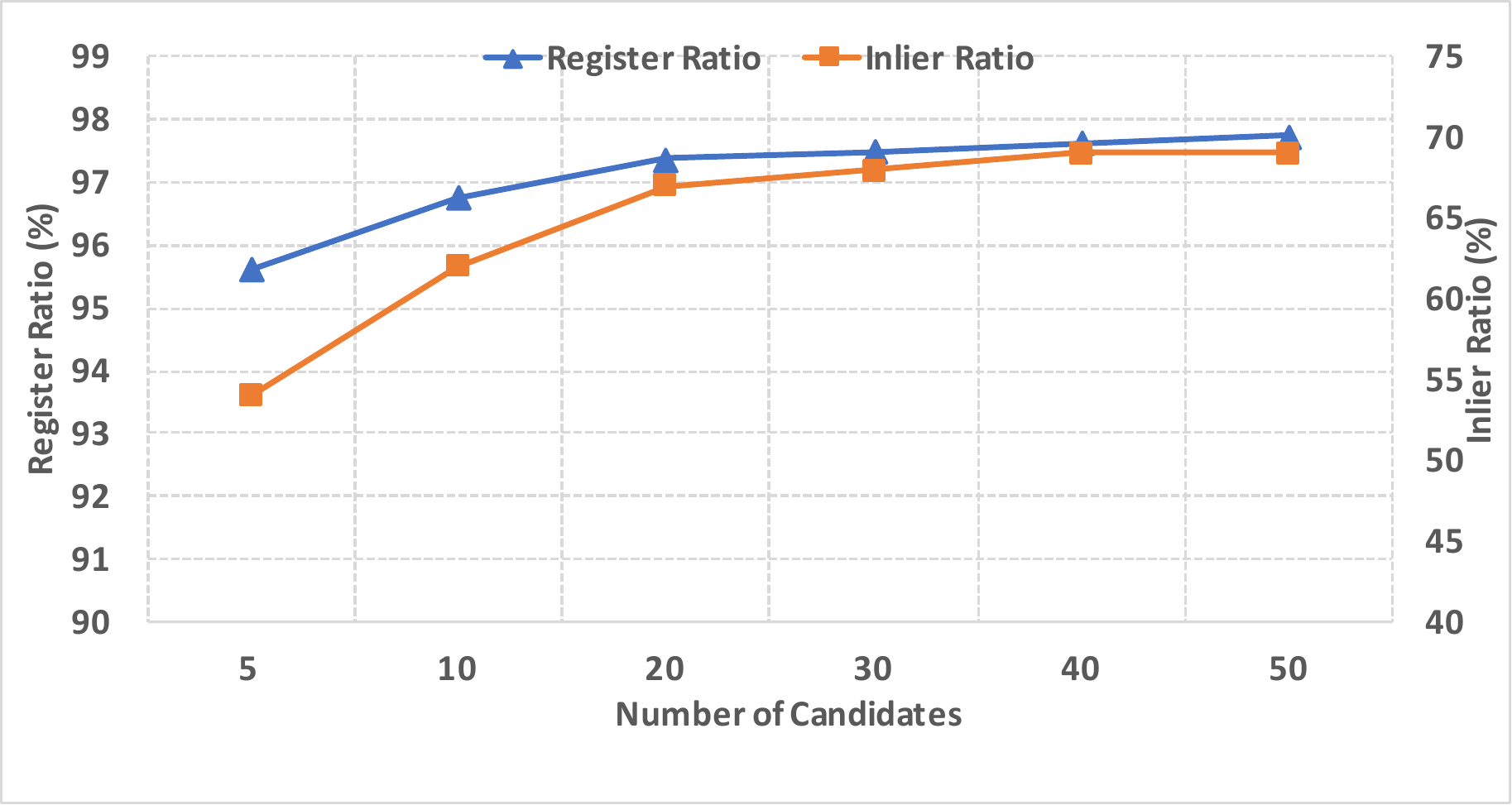}
	\caption{The registration rate and inliers ratio according to the number of candidates of $\mathbf{L}_R$.}
	\label{lR_validation}
\end{figure}

The second experiment to choose the good size of $|\mathbf{L}_R|$ output from refined search. Conditions like the first experiment, except we choose the best threshold $\nu_h = 0.8$ for precise search. We validate our method with a various number of candidates in $\mathbf{L}_R$. This experiment suggests that $|\mathbf{L}_R| = 40$ is a good option because increasing it does not significantly affect the registration rate and inliers ratio (Fig. \ref{lR_validation}).

\begin{figure}
	\centering
	\includegraphics[scale=0.41]{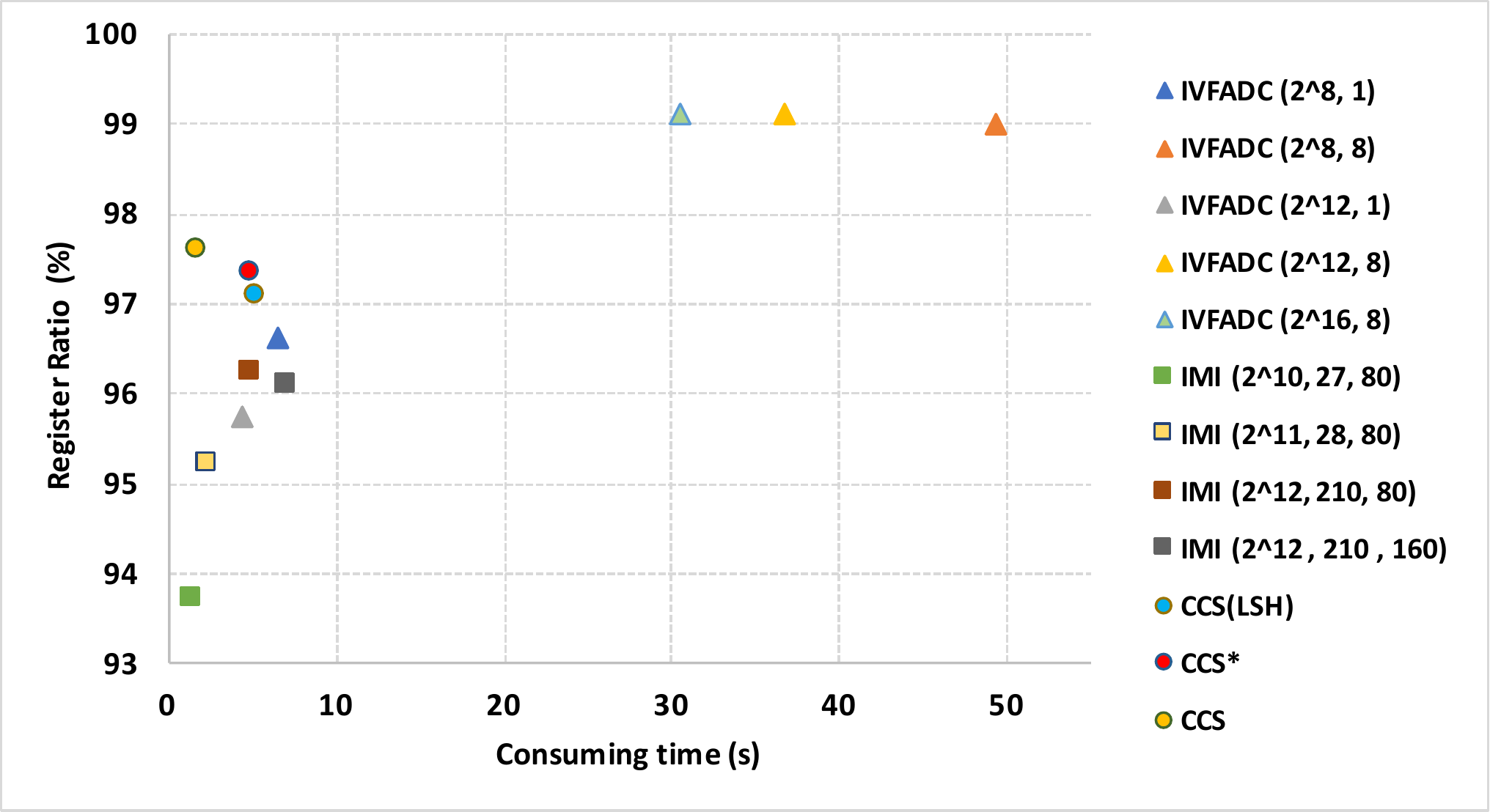}
	\caption{Comparison on indexing methods for PQ. Parameters for IVFADC ($K_{c},w$), IMI ($K_{c},w,|\mathbf{L}_R|$), CCS ($K_b = 2^{16}$, $|\mathbf{L}_R| = 40$). The version of our CCS is Setting 1, and `*' indicates our CCS ignoring the refined search.}
	\label{dubrovnik-pq-methods}
\end{figure}

In the third experiment on Dubrovnik dataset, we study the influence of different indexing procedures on accuracy and computation, by comparing our method against two well-known PQ-based indexing schemes. Similar to the second experiment, we compare our CCS to Inverted File (IVFADC) \cite{jegou-pami-2011} and Inverted Multi-Index (IMI) \cite{babenko-cvpr-2012}. We also compare to our own methods without refined search. We tune parameters of IVFADC and IMI for a fair comparison. Results in Fig. \ref{dubrovnik-pq-methods} demonstrate the efficiency of refined search, because removing this step slows CCS down about $\sim3\times$ (approximately three times), while obtaining similar registration rate. Although IVFADC with $w=8$ visited cells achieves highest performances with different sizes of sub-quantizers, it is too slow. Our method outperforms IVFADC (with $w=1$) in terms of execution time and registration rate. 
IMI registers more queries when the number of nearest neighbors $w$ or the length of its re-ranking list $\mathbf{L}_R$ (same meaning as ours) is increased. Yet it also increases processing time. Our registration rate is higher than IMI, while our running time is competitive. 
We try to replace ITQ by LSH \cite{charikar-acm-2002} based scheme \cite{cheng-cvpr-2014}. Results show that using our scheme with ITQ  is $\sim3\times$ faster than LSH scheme \cite{cheng-cvpr-2014}. This is consistent with the parameter of the number of candidates reported in Fig. \ref{dubrovnik-pq-methods}. Note that for all experiments above, we use 1-1 matchings and traditional RANSAC (Setting 1).

\begin{table*}[t]
\footnotesize
\centering
\caption{We compare our method to the state of the art on Dubrovnik dataset. Methods marked `+' reports only the processing time of outlier rejection/voting scheme, taken from original papers (ignoring the execution time of 2D-3D matching). Methods marked `*' report results after bundle adjustment. Here, CCS: Cascade Correspondence Search, P: Prioritizing, R$_{1-M}$: One-Many RANSAC.}
\begin{tabular}{|c|c|c|c|c|c|c|c|}
\hline 
\textbf{Method} & \textbf{\#reg. images} & \textbf{Median} & \multicolumn{2}{c|}{\textbf{Quartiles {[}m{]}}} & \multicolumn{2}{c|}{\textbf{\#images with error}} & \textbf{Time (s)}\tabularnewline
\hline 
 & & & 1st Quartile & 3st Quartile & $<$ 18.3m & $>$400m & \tabularnewline
\hline 
\hline 
Kd-tree & 795 & - & - & - & - & - & 3.4* \tabularnewline
\hline 
Li \etal \cite{li-eccv-2010} & 753 & 9.3 & 7.5 & 13.4 & 655 & - & \tabularnewline
\hline 
Sattler \etal \cite{sattler-iccv-2011} & 782.0 & 1.3 & 0.5 & 5.1 & 675 & 13 & 0.28 \tabularnewline
\hline 
Feng \etal \cite{feng-ip-2016} & 784.1 & - & - & - & - & - &\tabularnewline
\hline 
Sattler \etal \cite{sattler-bmvc-2012} & 786 & - & - & - & - & - & \tabularnewline
\hline 
Sattler \etal \cite{sattler-eccv-2012} & 795.9 & 1.4 & 0.4 & 5.3 & 704 & 9 & 0.25 \tabularnewline
\hline 
Sattler \etal \cite{sattler-pami-2016} & 797 & - & - & - & - & - & \tabularnewline
\hline 
Cao \etal \cite{cao-cvpr-2013} & 796 & - & - & - & - & - & \tabularnewline
\hline 
Camposeco \etal \cite{camposeco-cvpr-2017} & 793 & - & 0.81 & 6.27 & 720 & 13 & 3.2 \tabularnewline
\hline 
Zeisl \etal \cite{zeisl-iccv-215} & 798 & 1.69 & - & - & 725 & 2 & 3.78$^{+}$\tabularnewline
Zeisl \etal \cite{zeisl-iccv-215}* & 794 & \textbf{0.47} & - & - & 749 & 13 & - \tabularnewline
\hline 
Swarm \etal \cite{svarm-cvpr-2014} & 798 & 0.56 & - & - & \textbf{771} & 3 & 5.06$^{+}$\tabularnewline
\hline 
Li \etal \cite{li-eccv-2012} & \textbf{800} & - & - & - & - & - & \tabularnewline
\hline 
\textbf{Setting 1} (CCS) & 781 & 0.93 & 0.34 & 3.77 & 710 & 12 & 0.62 \tabularnewline
\hline 
\textbf{Setting 2} (CCS + R$_{1-M}$)& 796 & 0.89 & \textbf{0.31} & 3.67 & 717 & 17 & 0.62 \tabularnewline
\hline
\textbf{Setting 3} (CCS + P +  R$_{1-M}$) & 794 & 1.06 & 0.39 & 4.15 & 711 & 10 & \textbf{0.09} \tabularnewline
\hline
\end{tabular}
\label{dubrovnik-state-of-the-art}
\end{table*}

\begin{table}
\footnotesize
\centering
\caption{The number of registered images on Rome, Vienna, and Aachen datasets.}
\begin{tabular}{|c|c|c|c|}
\hline
\textbf{Method} & \textbf{Rome} & \textbf{Vienna} & \textbf{Aachen}\tabularnewline
\hline
\hline 
Kd-tree & 983 & 221 & 317 \tabularnewline
\hline 
Li \etal \cite{li-eccv-2010} & 924 & 204 & - \tabularnewline
\hline 
Sattler \etal \cite{sattler-pami-2016} & 990.5 & 221 & 318 \tabularnewline
\hline 
Cao \etal \cite{cao-cvpr-2013} & 997 & - & 329 \tabularnewline
\hline 
Sattler \etal \cite{sattler-bmvc-2012} & 984 & 227 & 327 \tabularnewline
\hline 
Feng \etal \cite{feng-ip-2016} & 979 & - & 298.5 \tabularnewline
\hline
Li \etal \cite{li-eccv-2012} & 997 & - & -\tabularnewline
\hline 
\textbf{Setting 2} (CCS + R$_{1-M}$) & 991 & \textbf{241} & \textbf{340} \tabularnewline
\hline 
\textbf{Setting 3} (CCS + P +  R$_{1-M}$) & 991 & 236 & 338 \tabularnewline
\hline 
\end{tabular}
\label{other-datasets-state-of-the-art}
\end{table}

\subsubsection{Pose estimation and prioritization}

\begin{figure*}[t]
	\centering
	\includegraphics[scale=0.15]{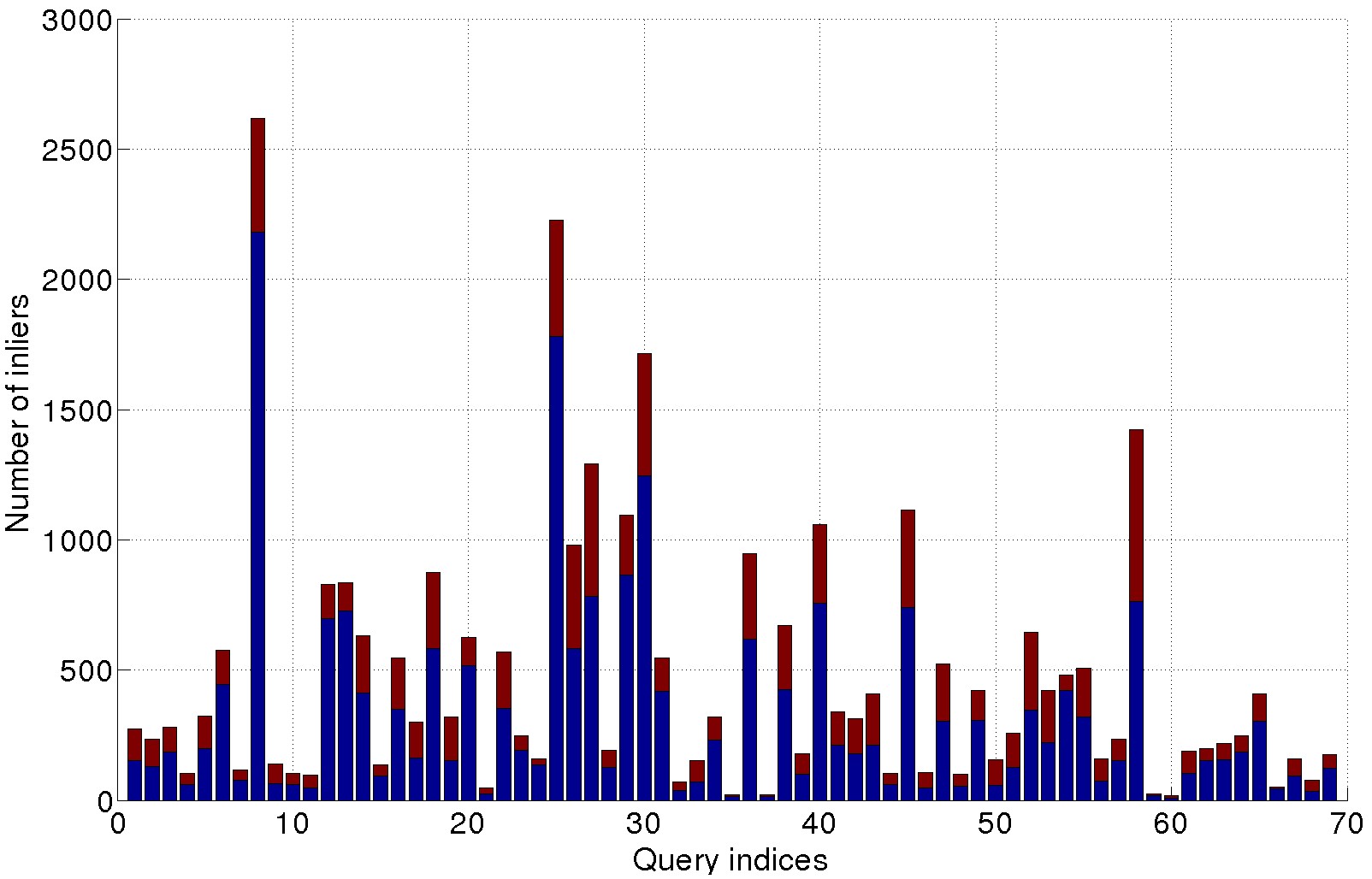}
	\includegraphics[scale=0.15]{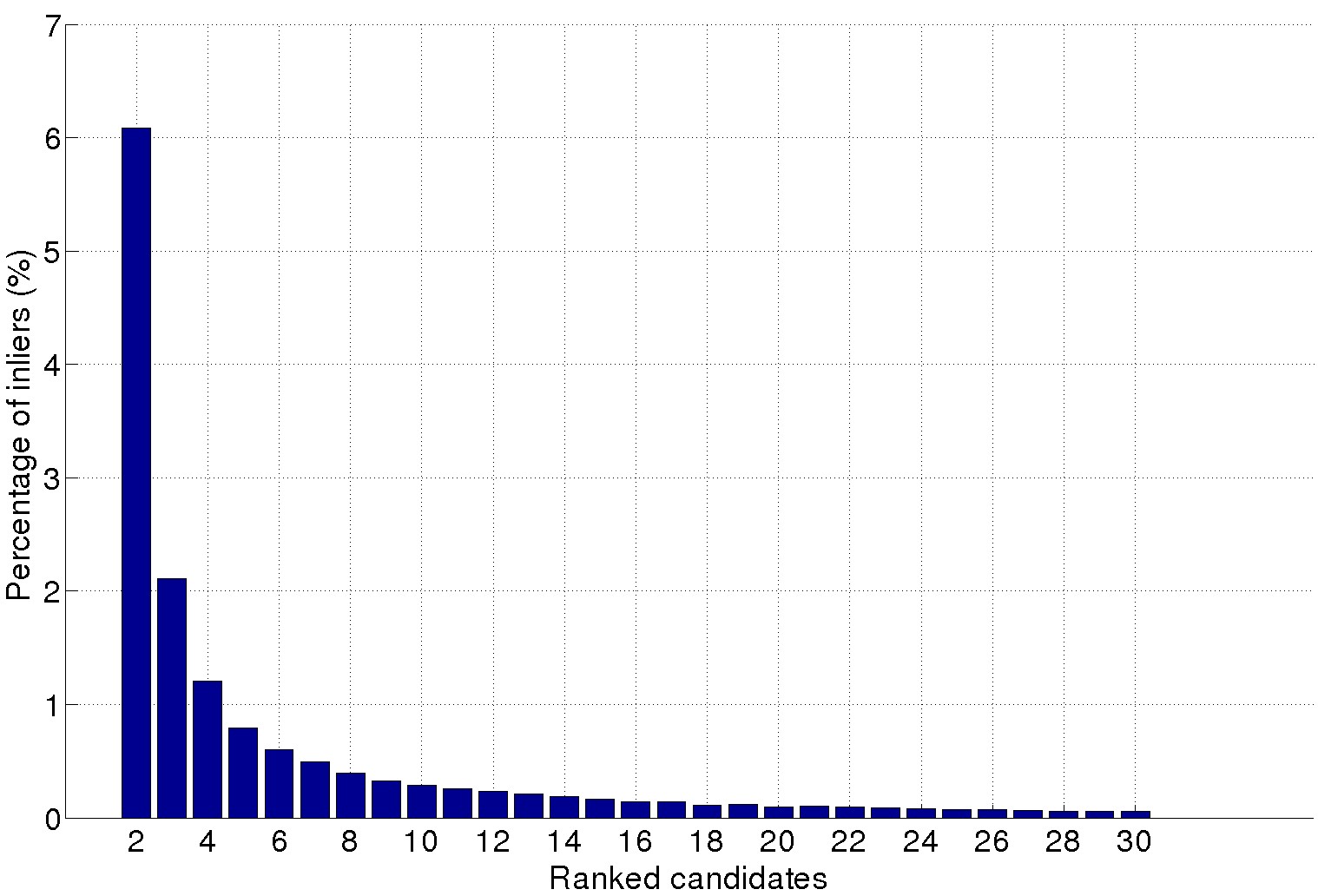}
	\includegraphics[scale=0.15]{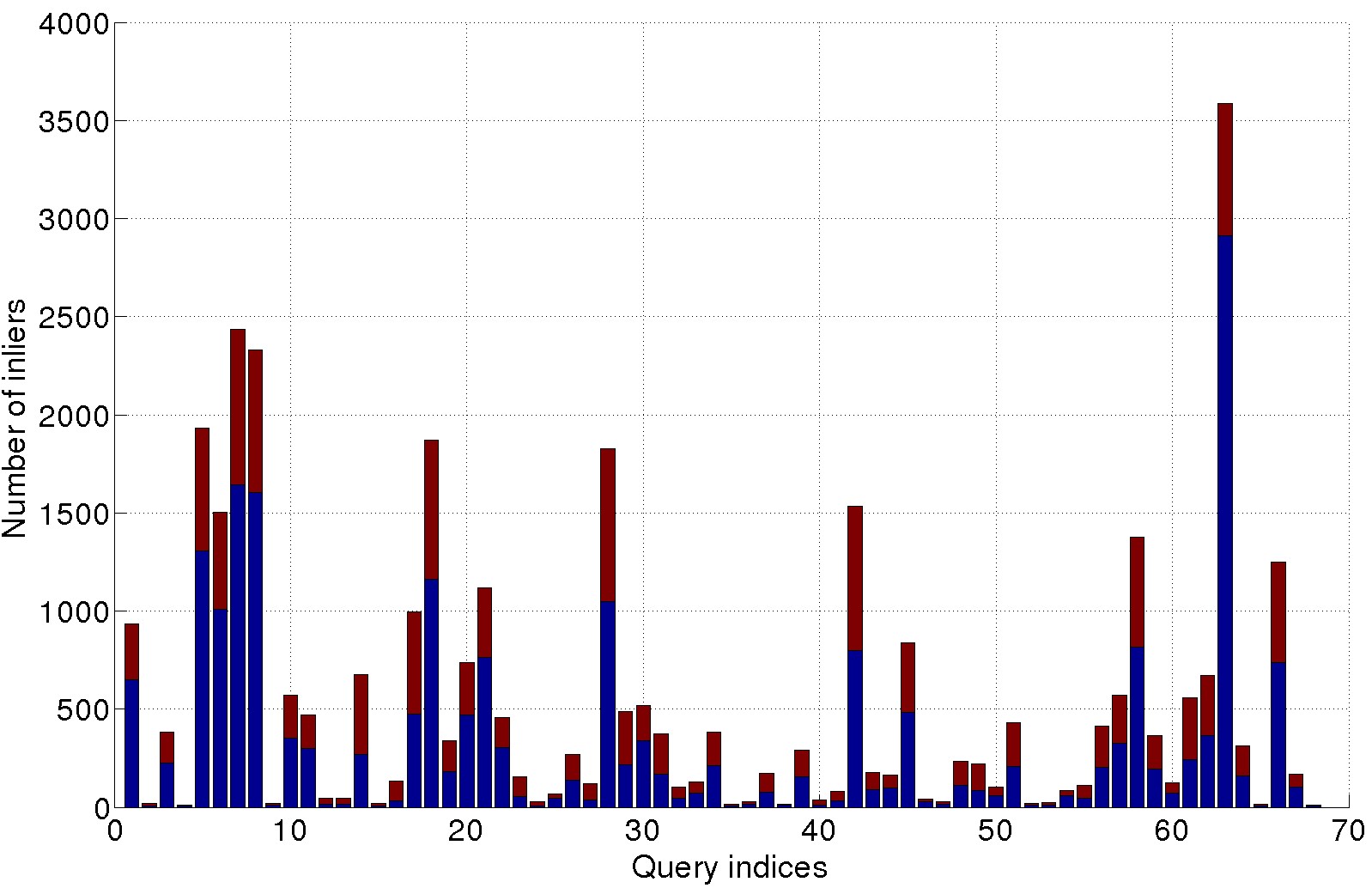}
	\includegraphics[scale=0.15]{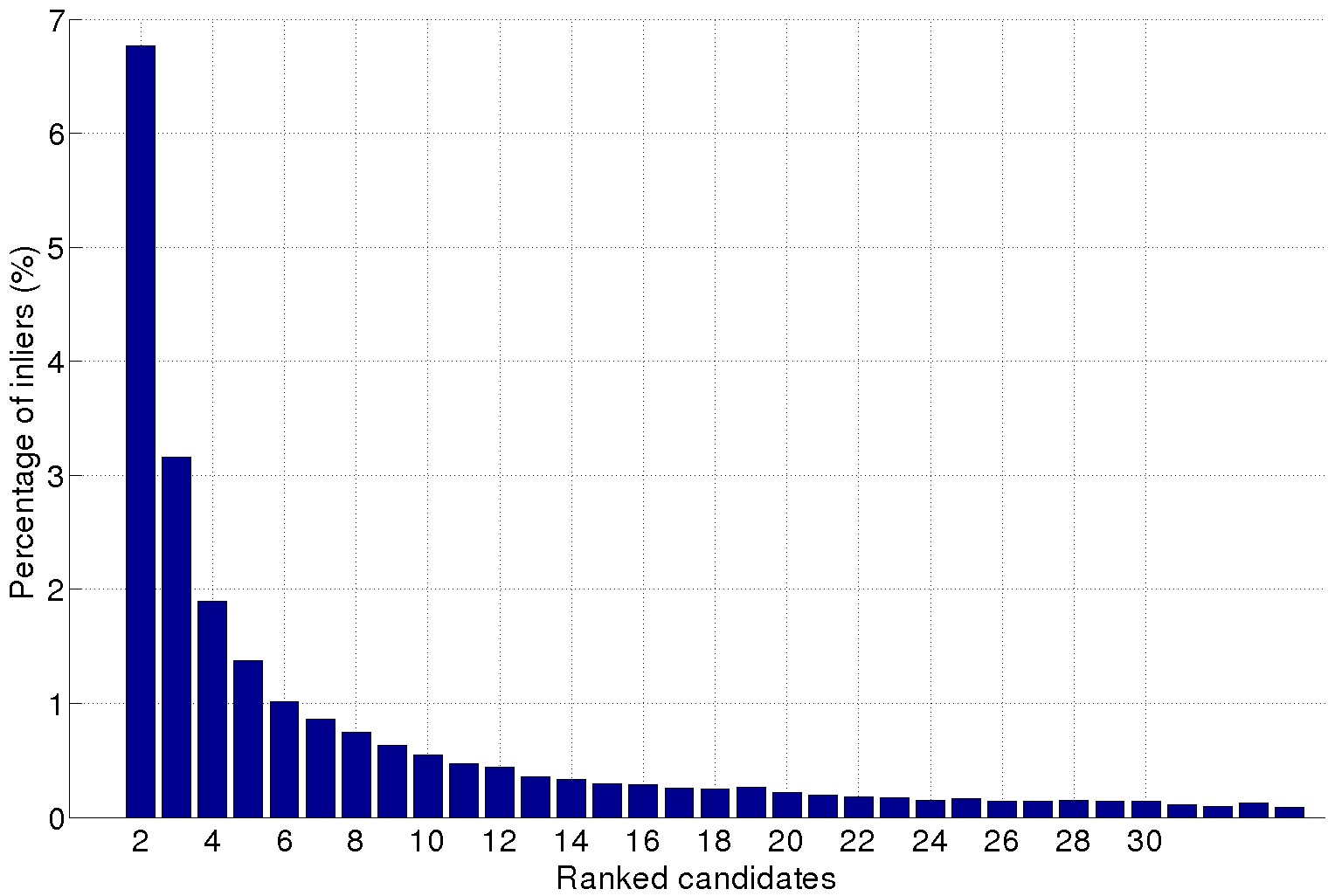}
	\caption{The contribution of inliers (on first 70 queries of Dubrovnik and Vienna) found by threshold $\nu_h = 0.8$ (blue), and additional inliers found by relaxed threshold $\nu = 0.9$ (red).}
	\label{inliers_ratio}
\end{figure*}

\begin{figure*}[t]
	\centering
	\includegraphics[scale=0.2]{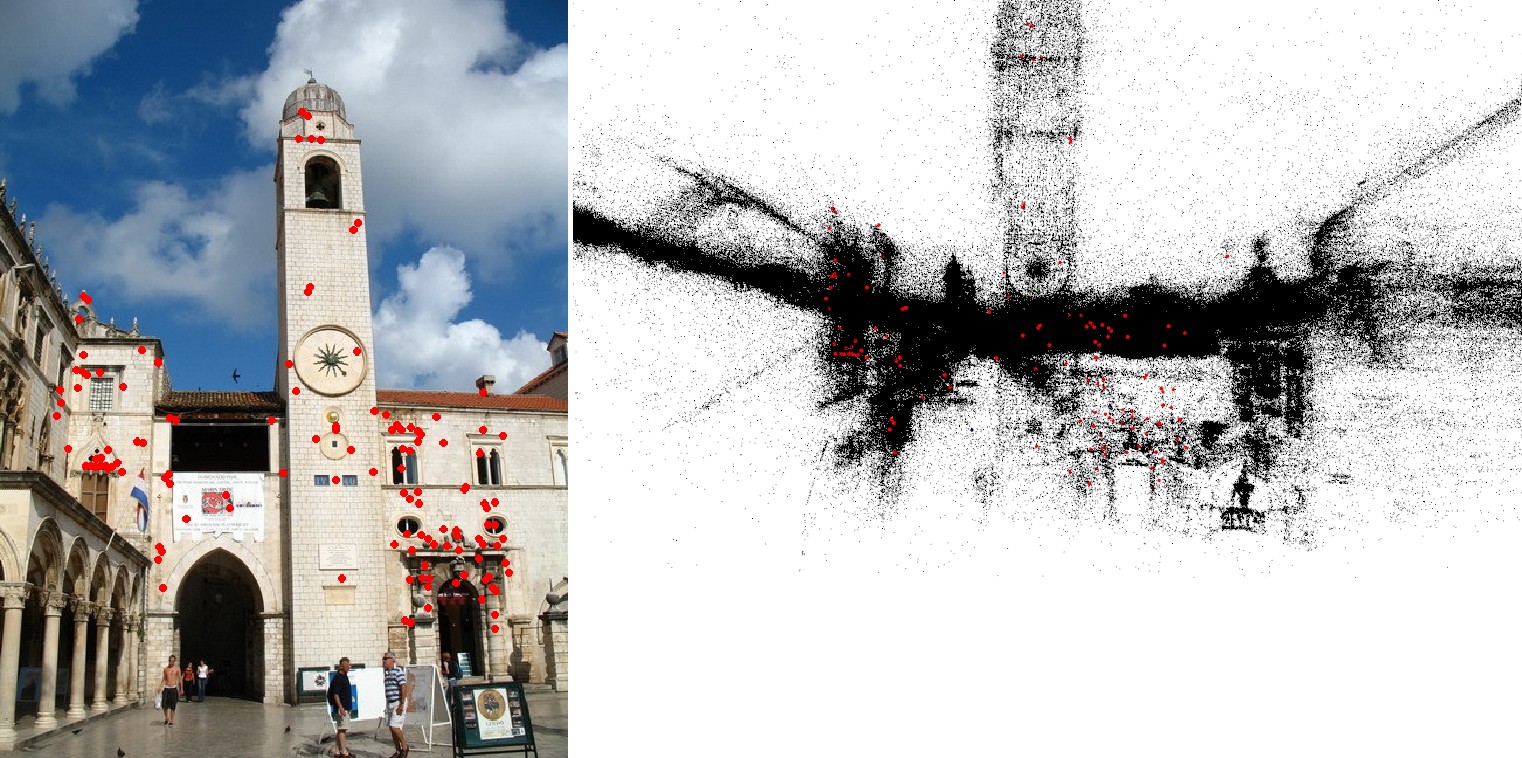}
	\includegraphics[scale=0.2]{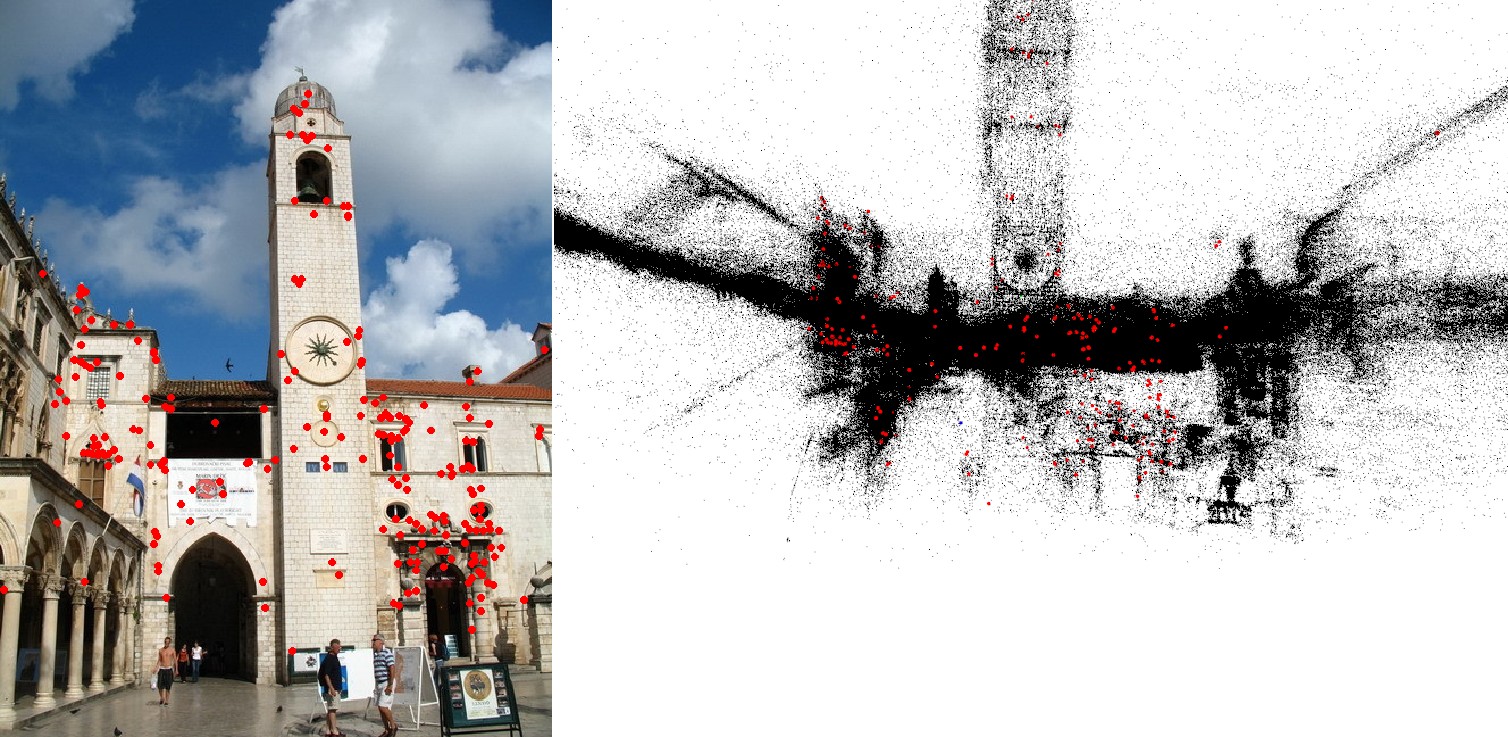}
	\caption{The left figure has 160 inliers found by our Setting 1 (with $\nu_h$), and the right figure has 278 inliers found by our Setting 2 (with the relaxed threshold $\nu$).}
	\label{visual_2d_3d_inliers}
\end{figure*}

In this section, we investigate the influence of our geometric verification (Setting 2), that combines cascade search and proposed RANSAC with a fixed number of 5000 iterations. We visualize the inliers found by our method on the Dubrovnik dataset to understand the impact of the ratio test. We adopt all candidates of $\mathbf{L}_R$, $M = |\mathbf{L}_R|$, in this experiment. Fig. \ref{inliers_ratio} The number of inliers per query on Dubrovnik (first row) and Vienna (second row) datasets. Left figures display the number of inliers (on first 70 queries of Dubrovnik/Vienna) found by threshold $\nu_h = 0.8$ (blue), and relaxed threshold $\nu = 0.9$ (red). Right figures are the percentage of number inliers contributed by candidates (from second order) in the list $\mathbf{L}_R$. On Vienna dataset, we increase approximately 100\% of inliers as using relaxed threshold contributes about nearly 48\% to the total of the number of inliers. The candidate list on Vienna dataset contributes a slightly higher number of inliers than that of Dubrovnik dataset. These explain why our method achieves better results on Vienna dataset. Fig. \ref{visual_2d_3d_inliers} shows inliers on one query example of Dubrovnik. For each query, the blue portion is the number of inliers found by the strict ratio $\nu_h$, and the red portion represent the additional ones found by the relaxed threshold $\nu$. On average, the relaxed threshold can increases about 65.4\% of inliers from the strict threshold, and contributes about 37.2\% to the total number of inliers found by our method (Setting 2). The right-hand-side of the first row is the average number of inliers contributed by 1-$M$ matchings (from the second rank). The 1-$M$ matches increase about average 15\% of the number of inliers from the strict threshold of 1-1, and about 7\% of the total. It means if we use $\nu$ threshold and 1-$M$ matchings, the method increases a significant number of inliers ($\geq 80\%$). We see on the right figure that lower ranked candidates $< 5$-th does not have significant impact on the total number of inliers; therefore to save on computation, we keep only $M = 5$ matchings after the precise search.

\begin{table*}
\footnotesize
\centering
\caption{The processing times of RANSAC and the registration times.}
\begin{tabular}{|c|c|c|c|}
\hline 
\textbf{Method} & \textbf{\#reg. images} & \textbf{RANSAC (s)} & \textbf{Reg. time (s)} \tabularnewline
\hline 
\hline 
Kd-tree & 795 & 0.001 & 3.4 \tabularnewline
\hline 
Sattler \etal \cite{sattler-iccv-2011} & 782.0 & 0.01 & 0.28 \tabularnewline
\hline 
Sattler \etal \cite{sattler-eccv-2012} & 795.9 & 0.01 & 0.25 \tabularnewline
\hline 
\textbf{Setting 3} (CCS + P +  R$_{1-M}$) & 794 &  0.20 & 0.29 \tabularnewline
\hline
\textbf{Setting 3$^+$} (CCS + P + Fast R$_{1-M}$) & 793 & 0.03 & \textbf{0.12} \tabularnewline
\hline
\end{tabular}
\label{dubrovnik-running-time-acs}
\end{table*}

\begin{table*}[t]
\footnotesize
\centering
\caption{The running times (including RANSAC) on Vienna and Aachen datasets.}
\begin{tabular}{|c|c|c|c|c|}
\hline
\textbf{Method} & \multicolumn{2}{c|}{\textbf{Vienna}} & \multicolumn{2}{c|}{\textbf{Aachen}}\tabularnewline
\hline
 & \textbf{\#reg. images} & \textbf{Reg. time (s)} & \textbf{\#reg. images} & \textbf{Reg. time (s)} \tabularnewline
\hline 
Sattler \etal \cite{sattler-iccv-2011} & 206.9 & 0.46 & - & - \tabularnewline
\hline 
Sattler \etal \cite{sattler-eccv-2012} & 220 & 0.27 & - & - \tabularnewline
\hline 
Sattler \etal \cite{sattler-pami-2016} & 221 & 0.17 & 318 & 0.12 \tabularnewline
\hline 
\textbf{Setting 3} (CCS + P +  R$_{1-M}$) & 236 & 0.35 & 338 & 0.28 \tabularnewline
\hline
\textbf{Setting 3$^+$} (CCS + P + Fast R$_{1-M}$) & 228 & \textbf{0.15} & 335 & \textbf{0.11} \tabularnewline
\hline

\end{tabular}
\label{other-running-time}
\end{table*}

Table \ref{dubrovnik-state-of-the-art} demonstrates the performance of our Setting 2 ($M = 5$). First, we see that our Setting 2 significantly outperforms our Setting 1 at both the number of registered images and errors. It confirms that using relaxed 1-$M$ candidates per query improves the performance. The registration rate and running time of Setting 2 is comparable to the state-of-the-art, however, its processing time can be further reduced by leveraging prioritizing scheme. We improve the cascade search speed with prioritized scheme (Setting 3). In the same Table \ref{dubrovnik-state-of-the-art}, Setting 3 obtains similar performance as the full search but it is about $\sim7\times$ faster. By using the prioritizing scheme, we achieve similar accuracy but with much faster matching speed than previous works. We also perform comparisons using other standard datasets (Table \ref{other-datasets-state-of-the-art}). Our Setting 3 outperforms the state-of-the-art methods in registration rate on Vienna and Aachen datasets. In addition to that, our proposed method is more efficient with regards to memory because of the use of compressed descriptors. Note that when possible, we run the 2D-3D matching methods on our machine and measure their running times (excluding RANSAC time). This shows the potential of using relaxed and 1-$M$ matches for better accuracy. However, our version of Setting 3 (fixed 5000 iterations) used in above experiments can be further improved in term of execution time.

We accelerate it by using pre-verification step (Setting 3$^+$). It preserves competitive accuracy, but $\sim20\times$ faster than RANSAC (5000 iterations) of Setting 3, as shown in Table \ref{dubrovnik-running-time-acs}. As a result, the total time of Setting 3 with our fast RANSAC is faster than Setting 3, and it needs a total of only 0.12(s) to successfully register one query. As compared to others, we outperform them in terms of registration rate and execution time on Vienna and Aachen datasets (Table \ref{other-running-time}). Our proposed RANSAC (Setting 3$^+$) executes as fast as classical RANSAC on the small set of correspondences, \eg 0.03(s) vs. 0.01(s) per Dubrovnik query in Table \ref{dubrovnik-running-time-acs}.

\begin{figure}
	\centering
	\includegraphics[scale=0.6]{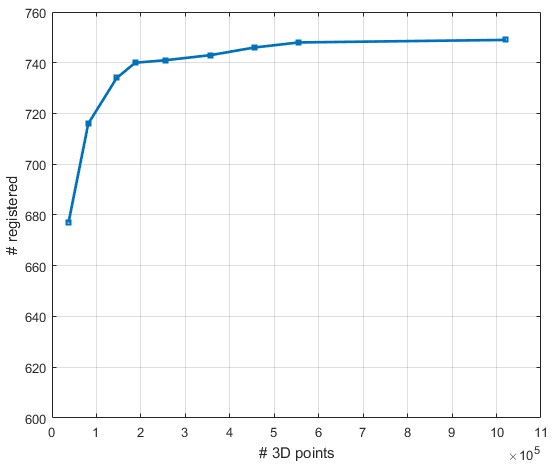}
	\caption{The number of 3D points of compressed Dubrovnik models and the registration rate of IVFADC method on the corresponding models.}
	\label{minimum_scene_experiment}
\end{figure}

As discussed in the next section, our model can reduce the memory requirements by the factor of about $\times 2$ from the original SIFT model. In this experiment, we compare our model to \cite{cao-cvpr-2014} for memory efficiency. We conduct this experiment on the Dubrovnik model ($1.8 \times 10^5$ 3D points) by using \cite{cao-cvpr-2014} to compress this model by certain factors and use IVFADC (which achieved the best registration rate among compared PQ methods, Fig. \ref{dubrovnik-pq-methods}) to obtain the registration on those compressed models. Fig. \ref{minimum_scene_experiment} shows that compressing Dubrovnik model to $10 \times 10^5$ 3D points (about $\times1.8$), the registration rate of IVFADC drops dramatically from 796 ($99.5\%$) to 750 ($93.75\%$). At similar compression factor (about $\times2$), our method can achieve about $97.3\%$ with Setting 1, and $99.25\%$ with our best Setting 3.
 
\subsection{Overall system}

\subsubsection{Google Street View (GSV) Dataset}
\label{gsv_model_design} 

\begin{figure}
	\centering
	\includegraphics[scale=0.158]{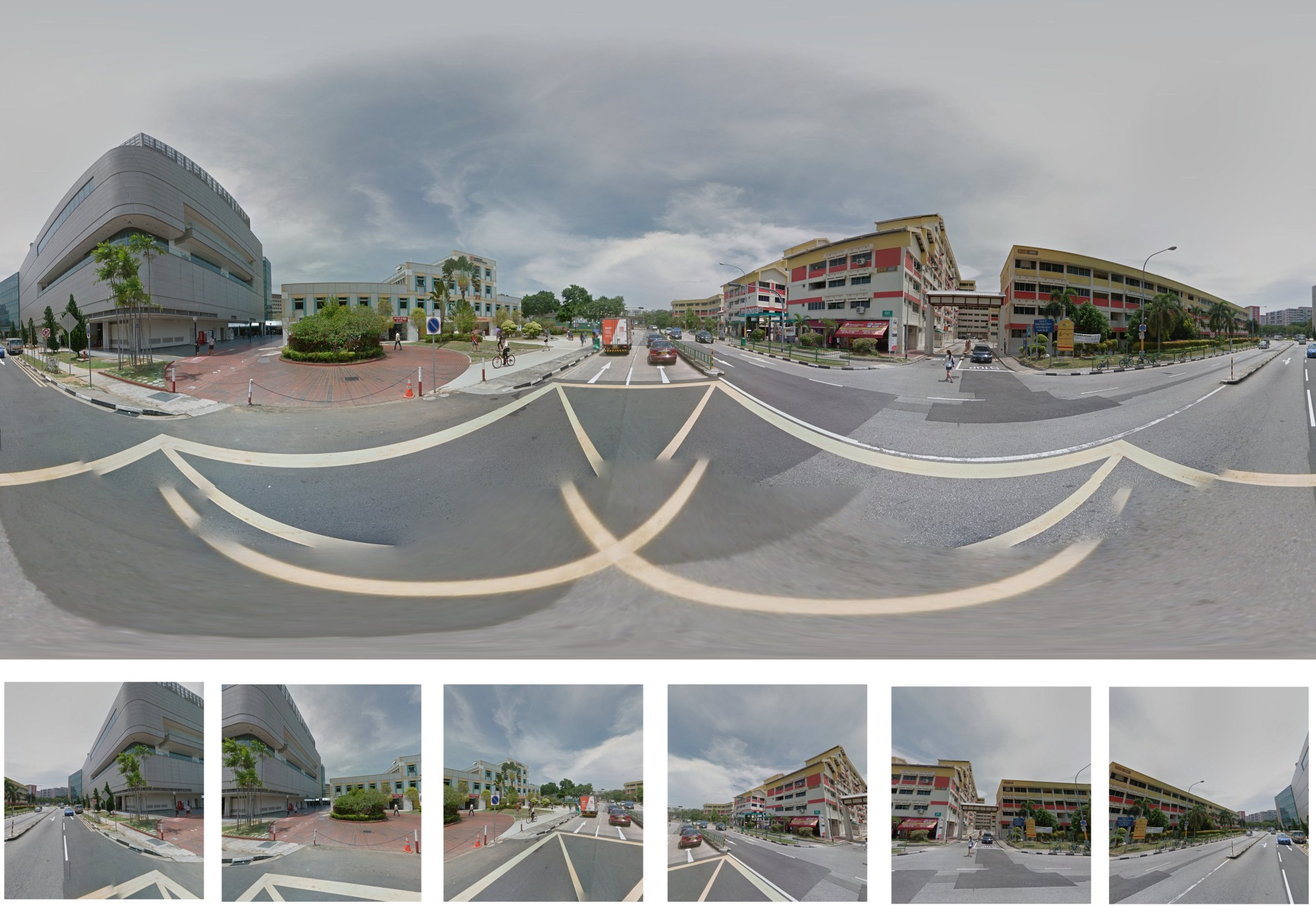}
	\caption{A panoramic image and its rectilinear views.}
	\label{panoramic_views}
\end{figure}

\begin{figure}
	\centering
	\includegraphics[scale=0.165]{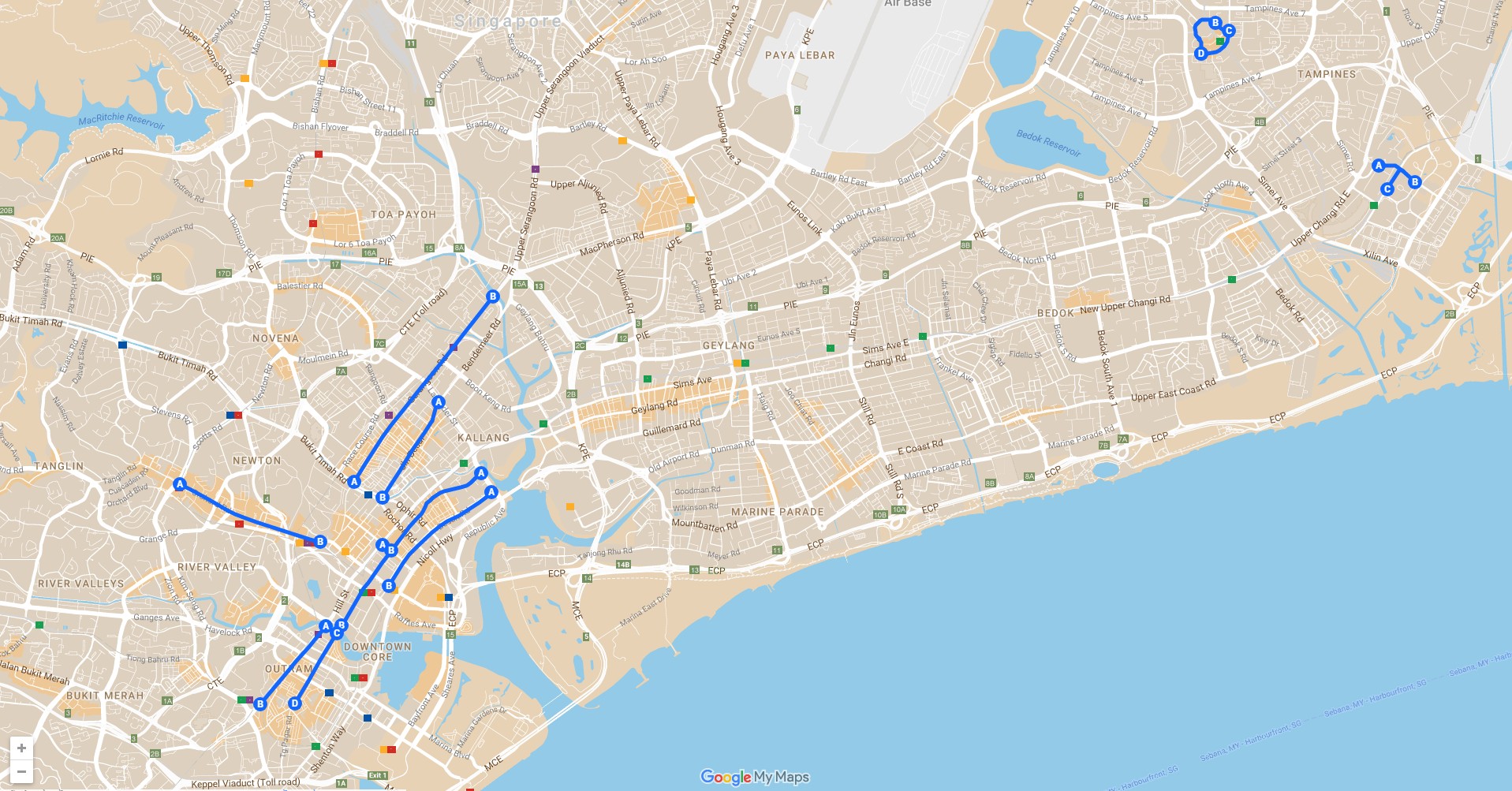}
	\caption{The coverage of our 200K dataset taking over about 15km road distance (roads marked by blue lines).}
	\label{fig_dataset_coverage}
\end{figure}

\begin{figure}
	\centering
	\includegraphics[scale=0.35]{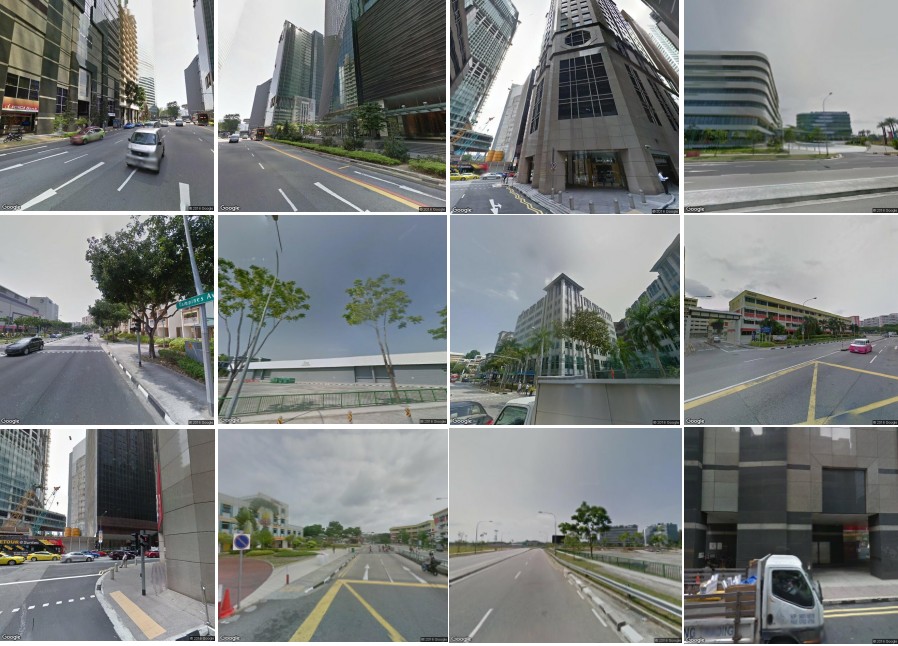}
	\caption{Examples of GSV images.}
	\label{fig_gsv_examples}
\end{figure}

\begin{figure}
	\centering
	\includegraphics[scale=0.42]{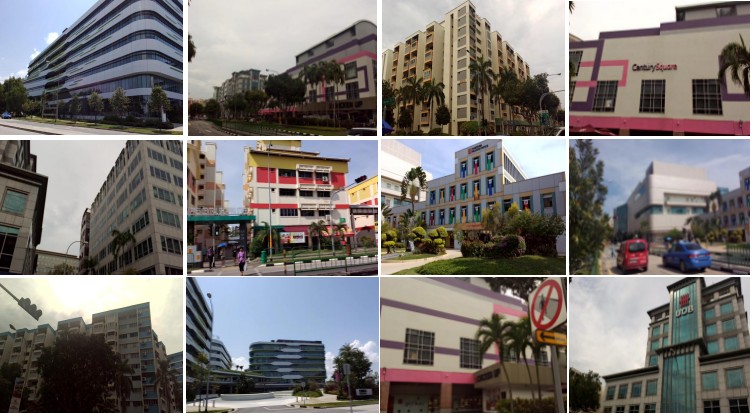}
	\caption{Examples of query images.}
	\label{fig_query_examples}
\end{figure}

We collect GSV images at a resolution of 640$\times$640 pixels. These images have exact GPS. We collect images that cover city regions in Singapore. At each Street View {\em place mark} (a spot on the street), the 360-degree spherical view is sampled by 20 rectilinear view images (18$^\circ$ interval between two consecutive side view images) at 3 different elevations (5$^\circ$, 15$^\circ$ and 30$^\circ$). Each rectilinear view has $90^\circ$ field-of-view and is considered a pinhole camera (Fig. \ref{panoramic_views}). Therefore, 60 images are sampled per placemark. The distance between two placemarks is about 10-12m. We also collect 576 query images with the accurate GPS ground-truth position. The GSV dataset for our training only supports scenes of the day, but the images are very distorted and challenging. Our mobile queries are collected with our own cameras under different conditions for a duration of several months. These conditions include
the morning, the afternoon with different lighting conditions and reflective phenomenon (building glass surfaces). See our dataset and query examples in Figures \ref{fig_gsv_examples} and \ref{fig_query_examples}. Our dataset covers about 15km road distance shown in Fig. \ref{fig_dataset_coverage}.

\begin{figure}
	\centering
	\includegraphics[scale=0.27]{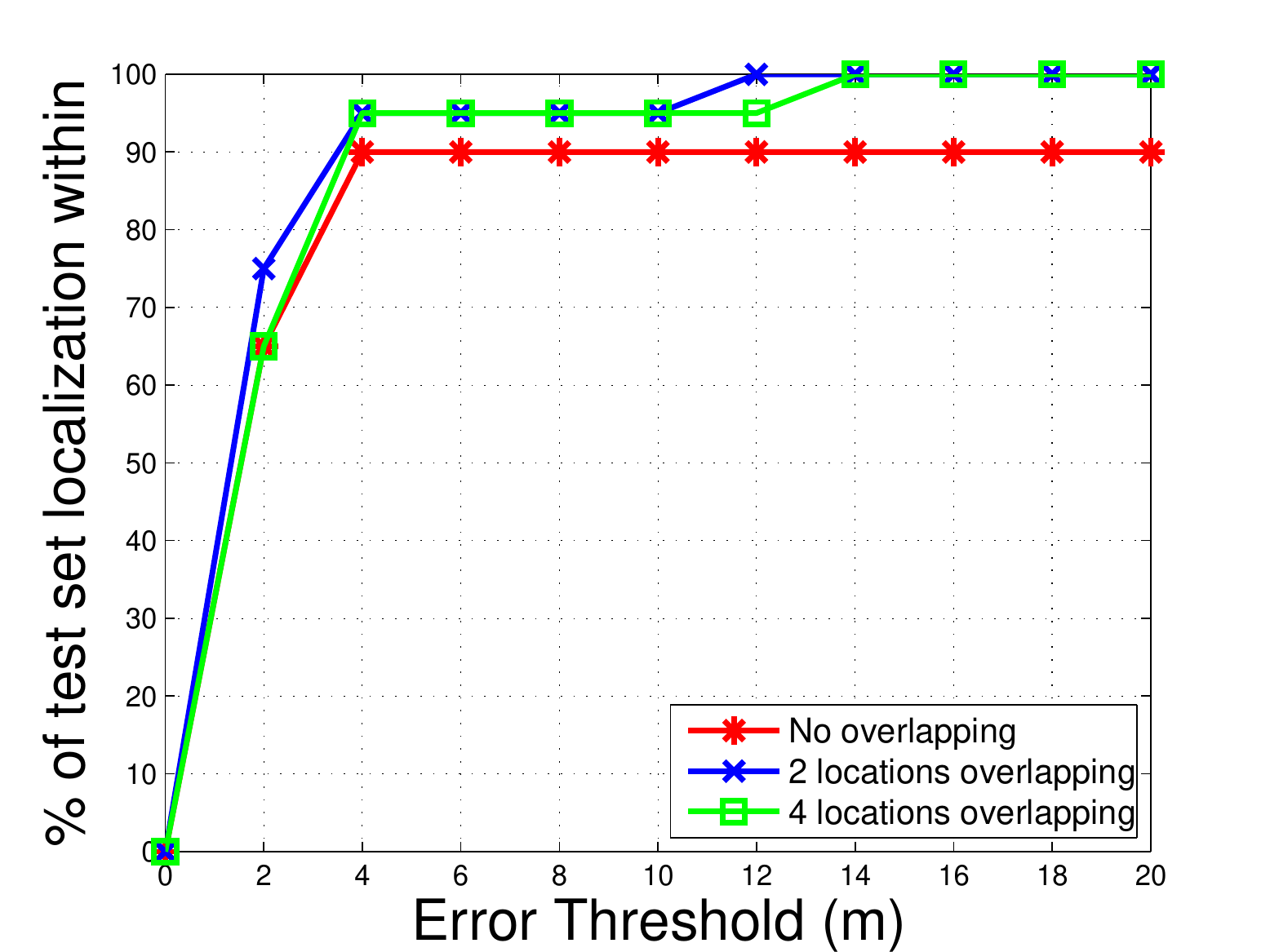}
	\includegraphics[scale=0.27]{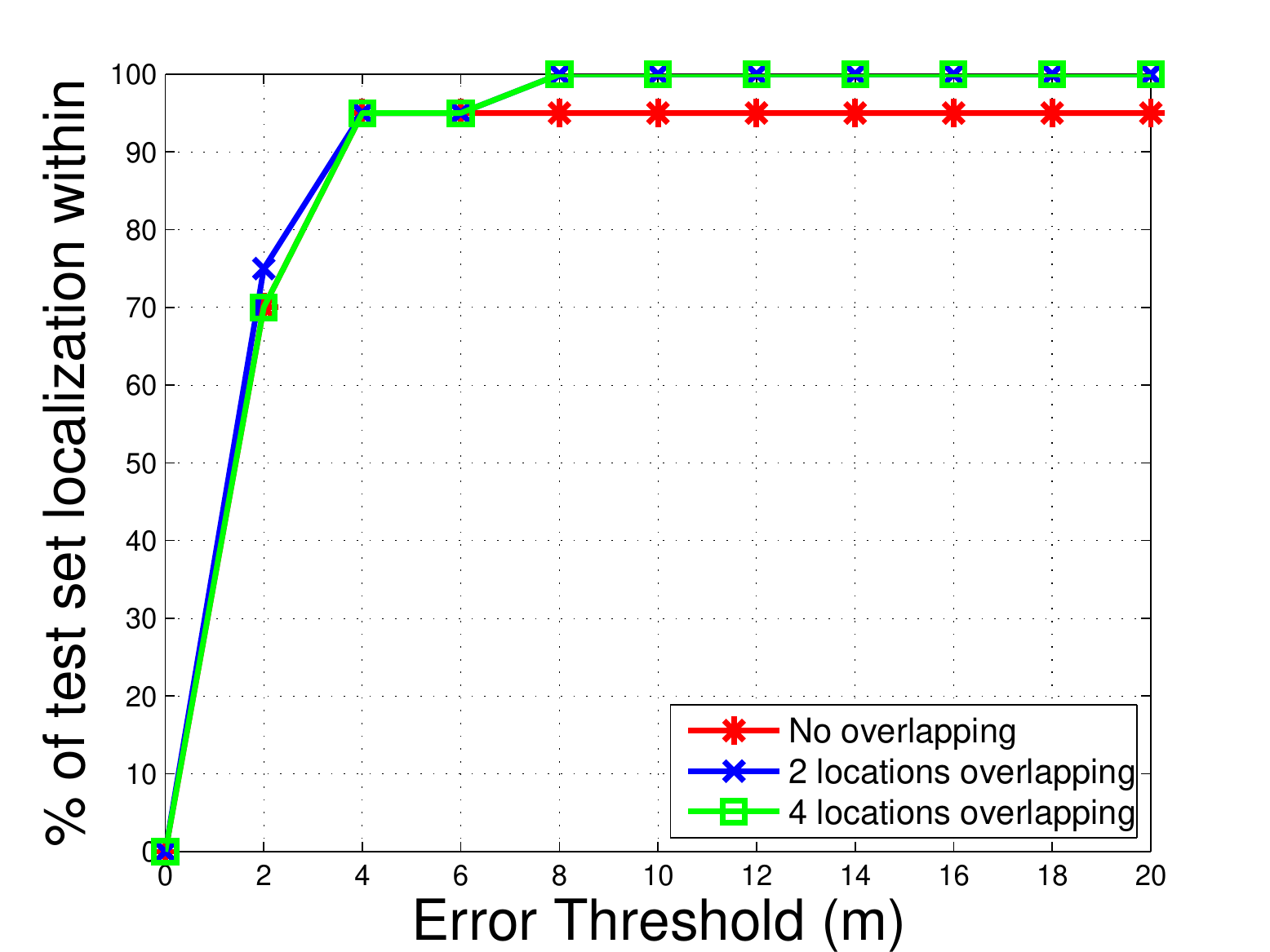}
	\caption{The number of overlapped placemarks of two segments, and its effects on the image-retrieval top list of 20 or 50. Two placemarks can ensure good localization accuracy. The retrieved list of 20 database images achieves good accuracy-speed trade-off.}
	\label{fig_overlaping_model}
\end{figure}

{\bf Overlapping of segments:}
We investigate overlapping between two consecutive segments. This is to ensure accurate localization for query images capturing buildings at the segment boundaries. We conducted an experiment to evaluate the localization accuracy at zero, two and four place marks overlapped. In this experiment, we used image retrieval to find the top 20 or 50 similar database images, given a query image. Results in Fig. \ref{fig_overlaping_model} suggest that with segments overlapped at two placemarks can ensure good localization accuracy. Note that the extent of overlapping is a trade-off between accuracy and storage. Besides, a retrieved list of 20 database images achieves good accuracy-speed trade-off.

\begin{table}
\footnotesize
\centering
\caption{The effect of segment size on the localization accuracy.}
\begin{tabular}{|c|c|c|}
\hline
\textbf{\#Place marks} & \textbf{\#Images} & \textbf{\% of \#queries with error $\leq$ 5m} \tabularnewline
\hline 
\hline 
 8-10  & 480-600 & 90\% \tabularnewline
 11-14 & 660-840   & 80\% \tabularnewline
 20-25 & 1200-1500 & 60\% \tabularnewline
\hline 
\end{tabular}
\label{fig_model_size_exp}
\end{table}

{\bf Coverage of each segment:} 
As the coverage (size) of each segment increases, the percentage of overlapped place marks decreases and hence storage (3D points) redundancy decreases.  However, the localization accuracy decreases as the segment coverage increases because there are more distracting 3D points in a 3D model. We conducted an experiment to determine an appropriate segment size:
we reconstructed a 3D model from a number of images: 480-600 images (8-10 placemarks), 660-840 images (11-14 placemarks), and 1200-1500 images (20-25 placemarks).  
We applied the state-of-the-art method, Active Correspondence Search (ACS) \cite{sattler-eccv-2012}, to compute the localization accuracy using the 3D model.
Table \ref{fig_model_size_exp} shows the results, which suggest that using segment with 8-10 placemarks achieves the best accuracy.  The localization accuracy degrades rapidly as we increase the segment coverage for GSV dataset. Therefore, in our system, we use 
8-10 consecutive GSV placemarks to define a segment. Although \cite{arth-ismar-2009} has also proposed to divide a scene into multiple segments, their design parameters have not been studied. Moreover, their design is not memory-efficient and covers only a small workspace area. It also requires prior additional sensor data, \eg GPS, WiFi to determine the search region. In addition, this work requires manual steps, \eg registering individual models into a single global coordinate. On the other hand, our models are automatically reconstructed or registered, and our system can localize entirely on a mobile device at a large scale.

In order to evaluate our on-device system, we consider the robustness of image retrieval on a large-scale dataset and the localization accuracy of an overall system on our GSV image collection.

\subsubsection{Image retrieval}

\begin{figure}
	\centering
	\includegraphics[scale=0.27]{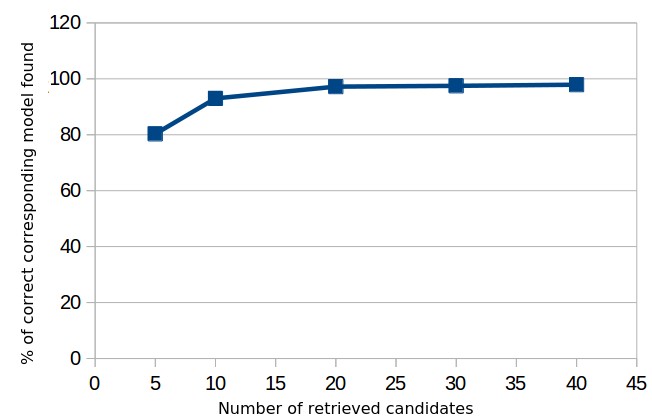}
	\includegraphics[scale=0.25]{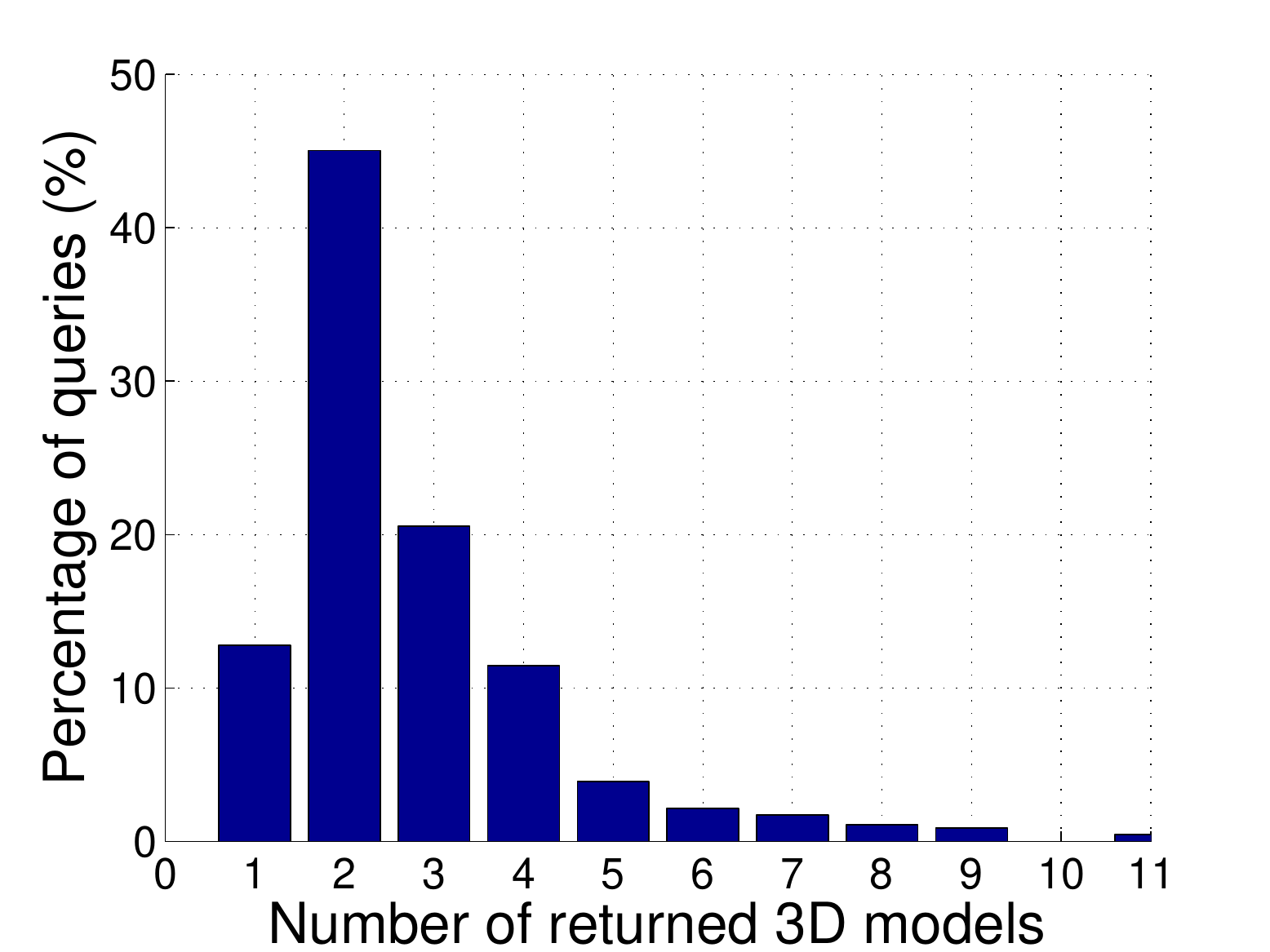}
	\caption{The accuracy of image retrieval and the histogram of the model number of threshold 20.}
	\label{image-retrieval-results}
\end{figure}

Image retrieval of our system finds the correct 3D models that a query is likely to belong. A query image is ``success", if $N_t$ top list of retrieval images match \textit{at least} one correct model. For ground-truth, we manually index our set of queries to their corresponding 3D models. It is important to investigate image retrieval performance, because it significantly affects the robustness of the overall system, especially with a large-scale dataset. We follow the parameters reported in T-embedding method \cite{jegou-cvpr-2014} as represented above and use sum-pooling. The goal is to determine the number of retrieved image $N_t$ that should be returned from image retrieval. $N_t$ has to balance between the accuracy and the number of models found. Fig. \ref{image-retrieval-results} shows that $N_t=20$ is an appropriate number. The horizontal axis represents the number of references resulted from image retrieval, and the vertical axis is the percentage of queries that found at least one correct model. The histogram of model numbers is visualized on the same figure. More than 80\% of queries found $\leq 4$ candidate models, therefore, we practically perform 2D-3D matching with the maximum number of four models if the list results in more than this number.

\subsubsection{Overall system localization}

\begin{figure}
	\centering
	\includegraphics[scale=0.6]{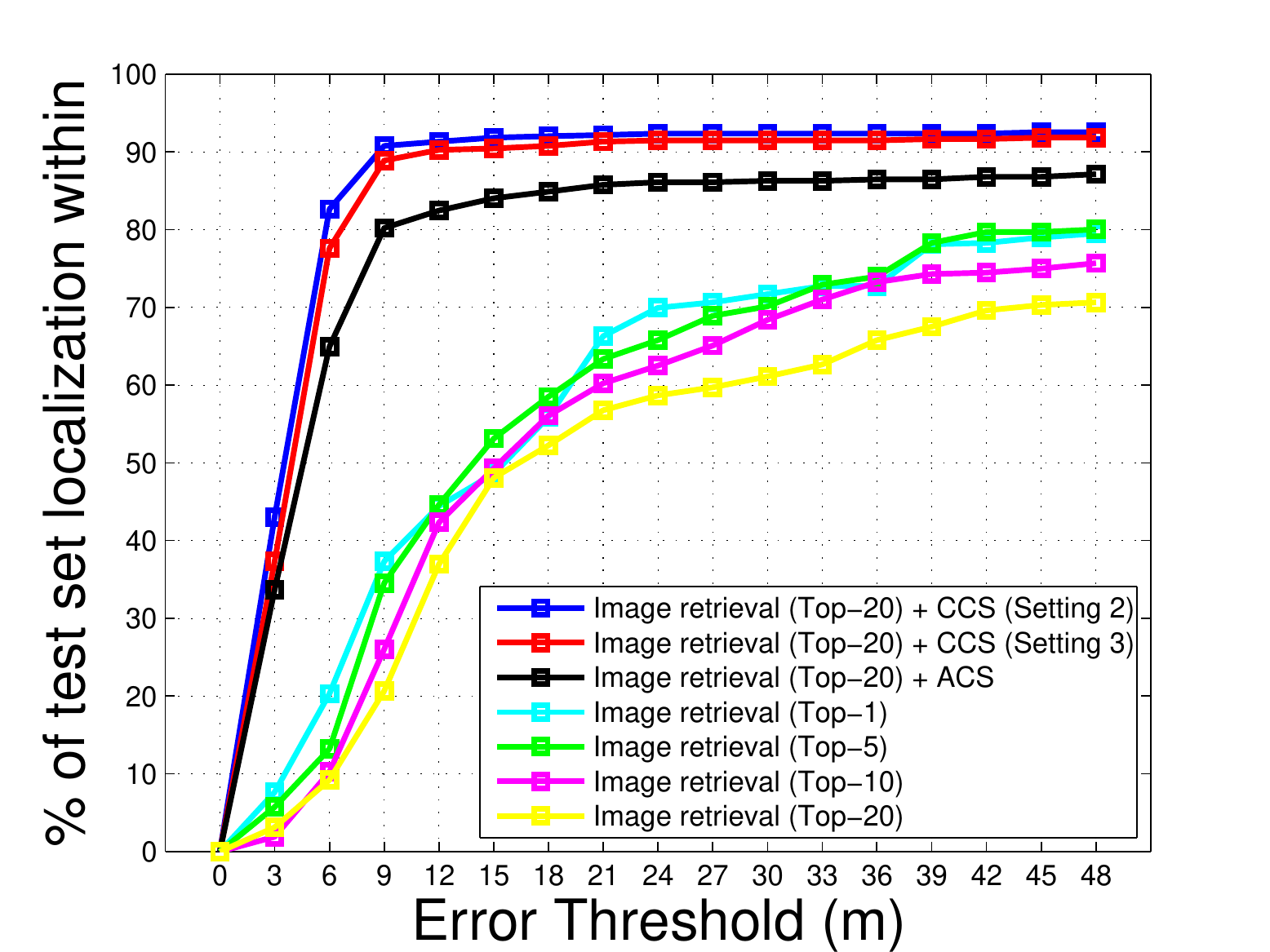}
	\caption{The performances of the overall system (image retrieval + 2D-3D correspondence search) tested on 576 query images. We also compare our 2D-3D matching method and ACS with the same image retrieval part. The performance of only image retrieval part also is reported, where top-k mean the GPS estimation is average of GPS of top k nearest images of the dataset according to the query image.}
	\label{overall-performance}
\end{figure}

In this experiment, the localization accuracy is measured by GPS distance between ground-truth and our estimation. The results are drawn in the form of Cumulative Error Distribution (CED) curve. The horizontal axis indicates the error threshold (in meters), and the vertical axis indicates the percentage of image numbers having lower or equal errors than the threshold. We compare correspondence search methods on the system: ACS \cite{sattler-eccv-2012} and our Setting 3$^+$. The same image retrieval component used for both that was trained on 227K images. Fig. \ref{overall-performance} presents real-world accuracies, \eg at the threshold of 9(m), about 90\% queries are well-localized for our Setting 3$^+$, slightly worse than our Setting 3 without our fast RANSAC, and about 80\% for ACS.
Our CCS uses compressed SIFT descriptors, which optimized better memory requirements than ACS but achieved better performance than ACS on our dataset. Note that the camera is calibrated in this experiment. About 10\% of images are completely failed ($\geq$ 50 (m)) due to image retrieval, the confusion of similar buildings, or reflection of building facades. Our proposed system achieves encouraging results using GSV images: the median error of our CCS (Setting 3$^+$) is about 3.75 (m), and 72\% of queries have errors less than 5 (m). In the same figure, we also evaluate the importance of using the localization part removing it from our system. The performance is drastically reduced without this part. The accuracy solely for the retrieval part is the average GPS of all retrieved images. It's worth noting that we may estimate better GPS by using some 2D-2D matching techniques between the query image and top retrieved ones. However, the disadvantage is that we need to store original images in the database and furthermore, the fusion of matching result is not simple.

\subsection{Memory Analysis and On-device computation}
\label{memory_and_time_complexity}

\subsubsection{Memory consumption of Image retrieval}

Our vocabulary size of T-embedding is $k_{temb}=32$, thus its T-embedding feature dimension is 4096. The fixed consumption of embedding and aggregating parameters is $67.14$ (MB). The indexing step of PQ needs approximately 129.44 (MB) to encode $N=227K$ images, where the number of sub-vectors is $g=256$ and the number of sub-quantizer per sub-vector is 256. The total memory is $129.44$ (MB), which can easily fit into modern devices with RAM $\geq 1GB$. When the size of a dataset is increased to 1M images, the total memory consumption of $327.34$ (MB) is still processable on the RAM.

\subsubsection{Memory consumption of 2D-3D correspondence search}

\begin{table}
\footnotesize
\centering
\caption{Memory requirements (\#bytes) for our model vs. original model.}
\begin{tabular}{|c|c|c|}
\hline
&  \textbf{Our model} & \textbf{Original model} \tabularnewline
\hline 
\hline 
\textbf{Look-up tables} & 8 $\times$ $2^{16}$ $\times$ 4 & - \tabularnewline
\hline 
\textbf{Point id} & $8 \times N_p \times 4 $ & $N_p \times 4$ \tabularnewline
\hline
\textbf{Point coordinates} & $N_p \times 12$ & $N_p \times 12$ \tabularnewline
\hline
\textbf{Descriptors} & $N_p \times (16 + 16)$ & $N_p \times 128$ \tabularnewline
\hline 
\textbf{Total memory} & $\approx N_p \times 76$ & $N_p \times 144$ \tabularnewline
\hline
\end{tabular}
\label{memory_requirements}
\end{table}

CCS (Setting 3$^+$) is implemented for mobile implementation as it is fast and requires less memory. The method requires 32 bytes (128-bit (16 bytes) hash code and 16 bytes PQ code) to encode a SIFT descriptor. Using $N_l=8$ look-up tables, each one comprised of $K_b=8\times2^{16}$ buckets. Each bucket needs a 4-byte pointer referring to one point-id list. Let $N_p$ be the point number of the 3D model if $N_p$ is large enough and small overhead memory can be ignored. $N_l$ tables refer to $N_l$ point-id with a total of $N_p$ points. One point-id can be represented by a 4-byte integer number. $N_p$ 3d point coordinates consume $N_p\times12$ bytes. Our model needs a total of $N_p\times76$ bytes, which is $\sim$2x more compressed than the original model (the 3D model of using SIFT descriptors) of $N_p\times144$ bytes shown in Table \ref{memory_requirements} (ignoring the indexing structures of other methods that may require more memory). Our 227K images of approximately 15km road distance coverage consume about 50MB of memory in total. We can extrapolate the numbers: It is feasible to extend to 1M images which can cover about 70 (km), while consuming less than 2GB memory.  We can extend the coverage further if storing 3D models on modern SD cards with large capacity. It is worth noting that the overall performance for such extensions would only affect  accuracy of image retrieval, not 2D-3D correspondence search, as we use scene partition and sub-models. Also, we have trained PQ sub-quantizers from the general dataset of 1M SIFT descriptors \cite{jegou-pami-2011}, which can be used for all models. The memory requirement for PQ sub-quantizers is: $256\times128\times4$ (bytes) $\approx$ 0.13 (MB).

Although our hashing scheme needs more memory as compared to two other PQ based schemes IVFADC and IMI that require 16-byte and 24-byte codes per 3D point, whose total memory is $N_p \times 32$ and $N_p \times 48$ (bytes) respectively, it is not critical as the size of the model is small enough to be loaded once on device memory; Furthermore, all models can be stored on an external device like SD cards. Our method is more efficient than two of these methods in terms of the trade-off between time complexity and accuracy reported on the Dubrovnik dataset.

\subsubsection{On-device running time}

Our system is implemented on Android device: Nvidia Tablet Shield K1, 2.2 GHz ARM Cortex A15 CPU with 2 GB RAM, NVIDIA® Tegra® K1 192 core Kepler GPU, 16GB storage. Our camera resolution is 1920$\times$1080. Table \ref{running-time} reports the running time for each individual steps: feature extraction, image retrieval, 2D-3D matching, and RANSAC. Since SIFT extraction is time-consuming, it is implemented using GPU. Image retrieval is also accelerated by GPU, whereas two other components used CPU. The processing time of image retrieval is acceptable and consistent with dataset size. Running time of 2D-3D matching is reported for only one model. On our dataset, the number of matches found is usually less than 100, hence the stopping early is not useful. In this case, our method obtains similar running time as ACS. In practice, a few models ($\leq 4$) are manipulated at a time and the latency of loading one model is low, about 0.04 (s). Therefore, it takes on average about 10 (s) in total to localize one query.  The localization and pose estimation parts are based on a single CPU core, the speed of our system can be further optimized/improved with multi-core CPU and GPU in future work. Note that we calculate the codebook size $K_{qc} = \frac{N_p}{10}$ when training ACS on our own models and other parameters using the same method reported in \cite{sattler-iccv-2015}.

\begin{table}
\footnotesize
\centering
\caption{Average running time for each individual step on our device.}
\begin{tabular}{|c|c|}
\hline 
\textbf{Step} & \textbf{Time (s)} \tabularnewline
\hline
\hline
Feature extraction (GPU) & 0.67 \tabularnewline
\hline 
Image retrieval (GPU) & 0.82 \tabularnewline
\hline 
2D-3D matching  & 0.55  \tabularnewline
\hline
Pose estimation & 1.15 \tabularnewline
\hline 
\end{tabular}
\label{running-time}
\end{table}


\section{Conclusion}

We present complete design of an entire on-device system for large-scale urban localization, by combining compact image retrieval and fast 2D-3D correspondence search. The proposed system is demonstrated via the dataset of 227K GSV images (with approximately 15km road segment). The scale of the system can be readily extended with our design. Experiment results show that our system can localize mobile queries with high accuracy. The processing time is less than 10s on a typical device. It demonstrates the potential of developing a practical city-scale localization system using the abundant GSV dataset.

We propose a compact and efficient 2D-3D correspondence search for localization by combining prioritized hashing technique and 1-M RANSAC. Our 1-M RANSAC can handle a large number of matches to achieve higher accuracy while maintaining the same execution time as traditional RANSAC. Our matching method requires $\sim$2x less memory footprint than using original models. Our matching method achieved competitive accuracy as compared to state-of-the-art methods on benchmark datasets, specifically we obtained the best performance of both processing time and registration rate on Aachen and Vienna datasets.

\bibliographystyle{IEEEtran}
\bibliography{biblio}
\end{document}